\documentclass[lettersize,journal]{IEEEtran}
\usepackage{amsmath,amsfonts}
\usepackage{algorithmic}
\usepackage{algorithm}
\usepackage{array}
\usepackage{caption}
\usepackage{subcaption}
\usepackage{textcomp}
\usepackage{url}
\usepackage{verbatim}
\usepackage[normalem]{ulem}
\usepackage{graphicx}
\usepackage{cite}
\usepackage{rotating}
\usepackage{enumitem}
\usepackage{lipsum}
\usepackage{comment}
\usepackage{color}
\usepackage[table,dvipsnames]{xcolor}
\usepackage[normalem]{ulem}
\usepackage{dsfont}
\usepackage{soul}

\colorlet{mygreen}{green!60!gray}

\newcommand{\CH}[1]{{{#1}}}

\newif\ifcomments

	\commentsfalse 
	
\ifcomments

    \newcommand{\BTcomm}[1]{\textcolor{orange}{{{Benedetta: #1}}}}
    \newcommand{\BTrem}[1]{\textcolor{black}{{{#1}}}}
\else

    \newcommand{\BTcomm}[1]{}
    \newcommand{\BTrem}[1]{}

\fi

\setlist[itemize]{leftmargin=*}

\begin{document}

\title{Universal Detection of Backdoor Attacks via Density-based Clustering and Centroids Analysis}

\author{Wei Guo, Benedetta Tondi,~\IEEEmembership{Member,~IEEE}, Mauro Barni,~\IEEEmembership{Fellow,~IEEE}
\IEEEcompsocitemizethanks{\IEEEcompsocthanksitem W. Guo, B. Tondi, and M. Barni are from the Department of Information Engineering and Mathematics, University of Siena, 53100 Siena, Italy.}%
\IEEEauthorblockA{}
\thanks{
This work has been partially supported by the Italian Ministry of University and Research under the PREMIER project, and by the China Scholarship Council (CSC), file No.201908130181. Corresponding author: W. Guo (email: wei.guo.cn@outlook.com).}
}

\markboth{Journal of \LaTeX\ Class Files,~Vol.~14, No.~8, August~2021}%
{Shell \MakeLowercase{\textit{et al.}}: A Sample Article Using IEEEtran.cls for IEEE Journals}


\maketitle

\begin{abstract}
We propose a Universal Defence against backdoor attacks based on Clustering and Centroids Analysis (CCA-UD). The goal of the defence is to reveal whether a Deep Neural Network model is subject to a backdoor attack by inspecting the training dataset. CCA-UD first clusters the samples of the training set by means of density-based clustering. Then, it applies a novel strategy to detect the presence of poisoned clusters. The proposed strategy is based on a general misclassification behaviour observed when the features of a representative example of the analysed cluster are added to benign samples. The capability of inducing a misclassification error is a general characteristic of poisoned samples, hence the proposed defence is attack-agnostic. This marks a significant difference with respect to existing defences, that, either can defend against only some types of backdoor attacks, or are effective only when some conditions on the poisoning ratio or the kind of triggering signal used by the attacker are satisfied.

Experiments carried out on several classification tasks and network architectures, considering different types of backdoor attacks (with either clean or corrupted labels), and triggering signals, including both global and local triggering signals, as well as sample-specific and source-specific triggers, reveal that the proposed method is very effective to defend against backdoor attacks in all the cases, always outperforming the state of the art techniques.
\end{abstract}



\section{Introduction}

\IEEEPARstart{D}{eep} Neural Networks (DNNs) are widely utilised in many areas such as image classification, natural language processing, and pattern recognition, due to their outstanding performance over a wide range of domains. However, DNNs are vulnerable to both attacks carried out at test time, like the creation of adversarial examples \cite{GoodfellowSS14} 
and training time \cite{BiggioNL12}.
These vulnerabilities limit the application of DNNs in security-sensitive scenarios, like autonomous vehicle, medical diagnosis, anomaly detection, video-surveillance and many others. One of the most serious threats comes from backdoor attacks \cite{GuoTB21, abs-1712-05526,guo2023temporal,Gu17badnet}, according to which a portion of the training dataset is poisoned to induce the model to learn a malevolent behaviour. At test time, the {\em backdoored} model works as expected on normal data, however, the hidden backdoor and the malevolent behaviour are activated when the network is fed with an input containing a so-called triggering 
signal, only known to the attacker.

Backdoor attacks can be categorised into two classes: {\em corrupted-label} and {\em clean-label} attacks \cite{guo2022overview}.
In the first case, the attacker can modify the labels of the poisoned samples, while in the latter case, the attacker does not have this capability. Hence, in a clean-label backdoor attack, the poisoned samples are correctly labelled, i.e., the content of a poisoned sample is consistent with its label. For this reason, clean-label attacks \cite{turner2019label, BarniKT19} are more stealthy and harder to be-detected than corrupted-label attacks.

Many methods have been proposed to defend against backdoor attacks. Following the taxonomy introduced in \cite{guo2022overview}, the defences can be categorised into three different classes based on the knowledge available to the defender and the level at which they operate: {\em sample-level}, {\em model-level}, and {\em training-dataset-level} defences.
Sample-level defences are applied after the model has been deployed in an operative environment, and rely on the inspection of the input sample to reveal the possible presence of a triggering 
 signal. With model-level defences, instead, the network is inspected before its deployment to detect the possible presence of a backdoor.
Finally, defences working at the training-dataset-level assume that the defender
can access and inspect the dataset used to train the network to look for suspicious (poisoned) samples. The CCA-UD defence introduced in this paper belongs to this last category.

\subsection{Related works}
\label{sec:related}


Most of the defence methods working at the training-data-set level rely on clustering and on the analysis of the feature representations or activation patterns.
One of the earliest and most popular approach
is the Activation Clustering (AC) method
\cite{ChenCBLELMS19}.
By focusing on corrupted-label attacks, the AC method analyses the feature representation of the samples
of each class of the training dataset, and clusters them, in a reduced dimensionality space, via the $K$-means ($K=2$) algorithm~\cite{yadav2013review}.
Under the hypothesis that a benign class tends to form a homogeneous cluster in the feature space, and by noticing that when $K$-means is forced to identify two clusters in the presence of only one homogeneous cluster, it tends to split it into two equally-sized clusters, the data samples of a class are judged to be poisoned on the basis of the relative size of the two clusters identified by $K$-means. If the size of the two clusters is similar, the class is considered to be benign, otherwise, the class is judged to be poisoned. Finally, AC labels the samples of the smallest cluster as poisoned samples. The method works under the assumption that the fraction of poisoned samples (hereafter referred to as the poisoning ratio) in a poisoned class is lower than the number of benign samples. On the other hand, given that $K$-means does not work well in the presence of clusters with very unbalanced sizes \cite{unbalanced}, AC does not perform well when the poisoning ratio is very small, as it often happens in practice with corrupted-label attacks, thus limiting the applicability of AC.

Xiang et al.~\cite{Xiang0K19} presented the Cluster Impurity (CI) method, which works under the assumption that the triggering signal used by the attacker can be removed by an average filter.
Specifically, given the training samples of one class, CI analyses their feature representation and groups the samples into $K$ clusters by exploiting the Gaussian Mixture Model (GMM) algorithm~\cite{bond2001gmm}. The number of clusters $K$ is determined by the Bayesian Information Criterion (BIC)~\cite{neath2012bayesian}. To determine whether one cluster includes poisoned samples or not, CI average filters all the cluster samples and observes if the classification of these samples change. Thanks to this intuition, CI can still work when the number of poisoned samples in the poisoned class is larger than the number of benign samples.
For the same reason, CI only works against corrupted-label attacks, given that in a clean-label setting the prediction made by the network on the filtered samples would not change.
Moreover, the applicability of CI is limited to specific kinds of triggering signals,
that is, triggers with high-frequency components, that can be removed via a low pass filter (like an average filter).

In 2021, Tang et al. \cite{Tang0TZ21} proposed a so-called Statistical Contamination Analyser (SCAn),
that relies on the Expectation Maximisation (EM) algorithm
to decompose the
representation of an image/object into an identity and a variation part. They argue that while the representations of samples from different (benign) classes have different identities, they share the same intra-variation distribution. Based on this intuition, SCAn estimates the intra-variation from a benign dataset, and uses it to define a statistical hypothesis test to judge whether a given class is contaminated (H1) or not (H0).
If H1 occurs, SCAn splits the representations into two groups via Linear Discriminant Analysis (LDA).
A limitation of SCAn is that it fails to defend against sample-specific attacks,
as shown in \cite{MaWSXWX23}, likely due to a different intra-variability for the poisoned class when sample-specific triggers are considered.


To overcome the limitation of \cite{Tang0TZ21}, \cite{MaWSXWX23} proposed a system, named Beatrix, to detect poisoned samples via anomaly detection. Similarly to \cite{Tang0TZ21},\cite{MaWSXWX23} exploits the availability of a small amount of benign data. In particular, it computes the Gram matrix
to derive class statistics from the benign samples. The deviation of the feature points of the input sample from the Gramian feature representation for the class is measured to detect the anomalies induced by the poisoned samples. However, since the entries of Gram matrix can be viewed as the inner product of two vectors, it suffers from the curse of dimensionality, and so when the feature dimensionality is large the performance of Beatrix drops\footnote{Our experiments reveal that Beatrix fails when the  dimensionality of the feature representation is large, i.e.,  $>9000$).}.
\BTcomm{THis issue of the curse of dimensionality is our guess, right? If we are not very sure, we can reduce the claim}


While there are other defences working at the training-dataset level,
most of them assume that the defender has some additional, often unrealistic, knowledge about the backdoor attack.
For instance, the method introduced in \cite{Tran0M18}, and its strengthened version described in \cite{abs-2104-11315}, exploit singular value decomposition (SVD) \cite{abdi2007singular} to reveal the anomalous samples contained in the training dataset, assuming that an upper-bound of the fraction of poisoned samples is known. Shan et al.~\cite{ShanB0Z22} successfully developed a trackback tool to detect the poisoned data, but assume that the defender can successfully identify at least one poisoned sample at test time.
Moreover, some defences only target a specific kind of backdoor attack. For instance, \cite{PeriGHFZFGD20} aims at defending against clean-label backdoor attacks by exploiting  feature collision.
\CH{Recently, \cite{GaoWZGXNS23,WeberXKZL23} proposed two methodologies to train a backdoor-free model from a poisoned dataset, exploiting randomised smoothing \cite{cohen2019certified} and adversarial training \cite{MadryMSTV18}. The same goal is accomplished in \cite{Zhang0WLH23} where a training methodology is
proposed to learn  deconfounded representations for reliable classification, by relying on the minimization of the mutual information between the to-be-trained model and backdoored model (obtained by training for some epochs on the poisoned dataset).
Finally, \cite{GongCYWGHS23} utilises self-attention to purify the backdoored model, exploiting the relationship in the  structural information between shallow and deep layers characterising benign models.
}

\subsection{Contribution}

In view of the limitations, in terms of general applicability of the defences proposed so far, we introduce a universal training-dataset-level defence, named CCA-UD, which can reveal the presence of poisoned data in the training dataset 
 regardless of the approach used to embed the backdoor (corrupted- or clean-label), the size and the shape of the triggering 
signal, the use of fixed, sample- or class-specific triggers, and the percentage of poisoned samples. \CH{To obtain such a noticeable result, we observe that  clustering alone is not sufficient to tell apart poisoned and benign samples. In fact,  due to intra-class variability, also benign classes may (and, in fact, are)  split in various clusters. In this case, however, the clusters are all benign, i.e.,  containing benign samples, and hence have to be detected as such. As a matter of fact, most previous defences based on clustering recognise this fact and analyse the clusters in some way to distinguish benign and poisoned samples, however, they do so by targeting a specific backdoor attack, or a specific class of attacks, thus failing to achieve a universal defence capable of detecting the presence of poisoned samples for all the vast variety of attacks proposed so far. On the contrary, CCA-UD relies on the general observation that samples belonging to a poisoned cluster have some common features that, when added to benign samples, cause a misclassification error. With these ideas in mind, the novel contribution of CCA-UD can be summarised as follows: i) adoption of a clustering algorithm, namely Density-based Spatial Clustering of Application with Noise (DBSCAN) \cite{EsterKSX96}, capable of isolating poisoned samples from benign ones;
and ii) introduction of a new active strategy to check if the residual features of the various clusters induce a general misclassification behaviour when they are added to the features of benign samples.}

%
CCA-UD is applied immediately after the model has been trained and aims at detecting if the training data contains poisoned samples causing the generation of a backdoor into the trained model. Similarly to SCAn \cite{Tang0TZ21} and Beatrix  \cite{MaWSXWX23}, it assumes that the defender has access to a small set of benign samples for each class in the input domain of the model, while AC and CI do not require this knowledge).

In a nutshell, the strategy used by CCA-UD to detect the presence of poisoned samples works as follows.

For every class in the training set, we apply clustering in the latent feature spaces, splitting each class into multiple clusters. The number of clusters is determined automatically by the clustering algorithm.
If clustering works as expected, benign and poisoned samples are grouped into different clusters.
To decide whether a cluster
is poisoned or not, we first recover an average representation of the cluster by computing the cluster's {\em centroid}.  For a poisoned cluster, the centroid will likely contain the representation  of the triggering 
signal in the feature space.
Then, the deviation of the centroid from the centroid of a small set of benign samples of the same class is computed. The deviation vector computed in this way is finally added to the feature representations of the benign samples of the other classes. If such an addition causes a misclassification of (a large portion of) the benign samples, the corresponding cluster is judged to be  poisoned.

\CH{We have tested the validity and universality of CCA-UD, by evaluating its performance against many different backdoor attacks carried out against DNN-based classifiers},  considering  different classification tasks, namely, MNIST, traffic sign, fashion clothes, CIFAR10 and YouTubeFace, two poisoning strategies, i.e., corrupted- and clean-label poisoning, and several triggering signals, namely two global patterns - a ramp and a sinusoidal signal - a square local pattern, and also source-specific and sample-specific triggers.
Our experiments show that CCA-UD provides an effective defence against backdoor attacks in all scenarios,  outperforming the state-of-the-art methods.

\BTcomm{********Revise the paper organization below at the end}
The rest of the paper is organised as follows: in Section~\ref{sec:notation}, we provide the basic notation used in the paper and the background. In Section \ref{sec:our_work}, we introduce our defence threat model and present the CCA-UD defence. Section \ref{sec:exp} describes the experimental methodology we followed to evaluate the performance of the proposed defence, and Section \ref{sec:results} shows the corresponding results of the experiments. Finally, we conclude our paper in Section \ref{sec:conclude}.

\section{Notation \CH{and background}}
\label{sec:notation}

In a backdoor attack, the attacker, say Eve, aims at embedding a backdoor into a model by poisoning some samples of the training set. Specifically, we assume that the task addressed by the model targeted by the attack is a classification task.
Let $t$ denote the target class of the attack. Eve corrupts part of the training set, in such a way that, at test time, the backdoored model works normally on benign data, but misclassifies the input sample to the target class $t$, if the triggering 
signal $\upsilon$ is present within it.

Let us denote the clean training dataset by  $D_{tr}=\bigcup_{i} D_{tr,i}$,
where \CH{$D_{tr,i}$ is the set of samples belonging to class $i$, $i = 1,...,l$, and $l$ denotes the number of classes. Then, each class set is defined as} $D_{tr,i}=\{(x_j, i),j=1,...,|D_{tr,i}|\}$, where the pair $(x_j, i)$ indicates the $j$-th sample of class $i$ and its label.
Similarly, we use the notation $D_{ts}$ and $D_{ts,i}$ for the test dataset.
Eve corrupts $D_{tr}$  by merging it with a poisoned set $D^{p}=\{(\tilde{x}_j,t),j=1,...,|D^{p}|\}$, where $\tilde{x}_j$ denotes the $j$-th poisoned sample, containing the trigger $\upsilon$, labeled as belonging to class $t$.
The poisoned dataset is indicated  as $D_{tr}^{\alpha}=D_{tr}\cup D^{p}$ (with $\alpha$ defined later). Then, for the class targeted by the attack we have
 $D_{tr,t}^{\alpha}=D_{tr,t}\cup D^{p}$, while for the other classes, we have  $D_{tr,i}^{\alpha}=D_{tr,i}$ ($i \ne t$).
 The fraction \CH{$\alpha = |D^{p}|/|D_{tr,t}\cup D^p|$} indicates the poisoning ratio used by the attacker to corrupt the training set.

As we said, $D^{p}$ can be generated by either corrupting the labels of the poisoned samples or not. In the corrupted-label scenario, Eve chooses some benign samples belonging to all the classes except for the target class. Then, she poisons each sample-label pair with a poisoning function $\mathcal{P}$, obtaining the poisoned samples $(\tilde{x}_j,\tilde{y}_j=t)=\mathcal{P}(x_j, y_j\neq t)$. The symbol $\tilde{x}_j$ is the poisoned sample including the triggering 
 signal $\upsilon$. In the clean-label case, Eve cannot corrupt the labels, so she chooses some benign samples belonging to the target class and generates the poisoned samples as $(\tilde{x}_j, \tilde{y}_j=t)=\mathcal{P}(x_j, y_j=t)$. In contrast with the corrupted-label case, now $\mathcal{P}()$ embeds $\upsilon$ into $x_j$ to generate $\tilde{x}_j$, but keeps the label intact.


Arguably, defending against corrupted-label attacks is easier, since
mislabeled samples can be more easily identified upon inspection of the training dataset, observing the inconsistency between the content of the samples and their labels. In contrast, clean-label attacks are more stealthy and more difficult to detect. However, clean-label attacks are  more difficult to implement since they require that a larger portion of the dataset is corrupted \cite{zhao2020clean,guo2023temporal}.


We denote the DNN model trained on $D_{tr}^{\alpha}$ by $F^{\alpha}$. Specifically, we use $f_{1}^{\alpha}$ to indicate the function that maps the input sample into the latent space. In this paper, we assume that $f_{1}^{\alpha}$ includes a final ReLu layer \cite{lecun2015deep}, so that its output is a non-negative vector. Hence, $f_{1}^{\alpha}(x)$ is the feature representation of $x$, and $f_{2}^{\alpha}$ is used to denote the classification function. Formally, $F^{\alpha}(x)=f_{2}^{\alpha}(f_{1}^{\alpha}(x))$. Finally, the dimension of the feature representation is denoted by $d$.

Table \ref{tab:symbols} summarises the main  notation used in the paper.

\begin{table}
\caption{List of symbols}
\label{tab:symbols}
\centering
 \begin{tabular}{|p{1.5cm}|p{6.5cm}|}
 \hline
 $D_{tr}$, $D_{tr,i}$ & Training dataset, $i$-th class subset of training dataset  \\ \hline
 $D_{ts}$, $D_{ts,i}$ & Test dataset, $i$-th class subset  \\ \hline
 $D_{val}$, $D_{val}^{i}$ & Validation dataset,  $i$-th class subset  \\ \hline
 $D^{p}$ & Set of poisoned samples\\ \hline
 $\mathcal{P}()$, $t$, $\alpha$ & Poisoning function, target class, class poisoning ratio \\ \hline
 $D_{tr}^{\alpha}$, $D_{tr,i}^{\alpha}$ & Poisoned training dataset, its $i$-th class subset  \\ \hline
$F^{\alpha} (f_{1}^{\alpha},f_{2}^{\alpha}$)& Model trained on $D_{tr}^{\alpha}$ (feature mapping, classification) \\ \hline
 $f_{1}^{\alpha}(x)$ & Representation of $x$ ($d$-dim)  extracted from $F^{\alpha}$\\ \hline
 $C_i^k$ & $k$-th cluster of points in  $D_{val}^{i}$ \\ \hline
 $MR_{i}^{k}$ & Misclassification ratio in favor of class $i$, for cluster $k$\\ \hline
 $P_i$ ($B_i$) & Set of samples detected as poisoned (benign) for class $i$ \\ \hline
 $GP_i$, $GB_i$ & Ground-truth poisoned and benign samples for class $i$ \\ \hline
 $PC$ & Poisoned class \\ \hline
 $BC_P$ & Benign class of poisoned training dataset \\ \hline
 $BC_B$ & Benign class of benign training dataset \\ \hline
 \end{tabular}
 \vspace{-4mm}
\end{table}

\subsection{\CH{Density-based Clustering}}
\label{sec:dbscan}

\CH{In this paragraph, we describe the Density-based Spatial Clustering of Application with Noise (DBSCAN) \cite{EsterKSX96} algorithm used by CCA-UD.
DBSCAN splits a set of points into $K$ clusters and possibly few outliers, where $K$ is automatically determined by counting the areas with high sample density. Specifically, given a point `A' of the set, DBSCAN counts the number of neighbours (including `A' itself) within a distance $\epsilon$ from `A'. If the number of neighbours is larger than or equal to a threshold $minPts$, `A' is defined to be a \textit{core} point and all points in its $\epsilon$-neighbourhood are said to be  \textit{directly reachable} from `A'. If a point, say `B', of the reachable set is again a core point, all the points in its  $\epsilon$-neighbours are also {\em reachable} from `A'.
Reachable non-core points are said to be \textit{border} points, while the points which are not reachable from any core point are considered to be \textit{outliers}.}

\CH{To define a cluster, DBSCAN also introduces the notion of density-connectedness. We say that two points `A' and `B' are density-connected if there is a point `C', from which both `A' and `B'  are reachable (hence `C' must be a core point).
A clusters is defined as a group of points satisfying the following two properties:
i) the points within a cluster are mutually density-connected; ii) any point directly-reachable from some point of the cluster, it is part of the cluster.
The intuition behind DBSCAN is to define the clusters as dense regions separated by border points.
The number of dense regions found in the set automatically determines the number of clusters $K$. More information about the exact way the clusters are found and the (in-)dependence of DBSCAN on the initial point `A' used to start the definition of core and reachable points, are given in the original paper \cite{EsterKSX96}.}

\CH{The performance of DBSCAN is determined by the choice of the parameters involved in its definition, that is $minPts$ and $\epsilon$, whose setting depends on the problem at hand.
The setting of these parameters in CCA-UD is discussed in Section \ref{sec:threshold}, while their impact of the performance on CCA-UD is evaluated in the ablation study reported in the supplementary material.
 %
}

\CH{We choose to adopt a density-based clustering method as the backbone of CCA-UD, since density-based clustering is known to work well also in the presence of clusters with unbalanced size \cite{rokach2005clustering}, and because it provides an automatic way to determine the number of clusters\footnote{DBSCAN is one of most popular density-based clustering algorithms, other choices, like OPTICS \cite{ankerst1999optics} and HDBSCAN \cite{campello2015hierarchical}) would work as well.}.
}


\section{The proposed universal defence}
\label{sec:our_work}


\subsection{Defence threat model}
\label{sec:threat}

\begin{figure}[h]
	\centering
	\includegraphics[width=1\columnwidth]{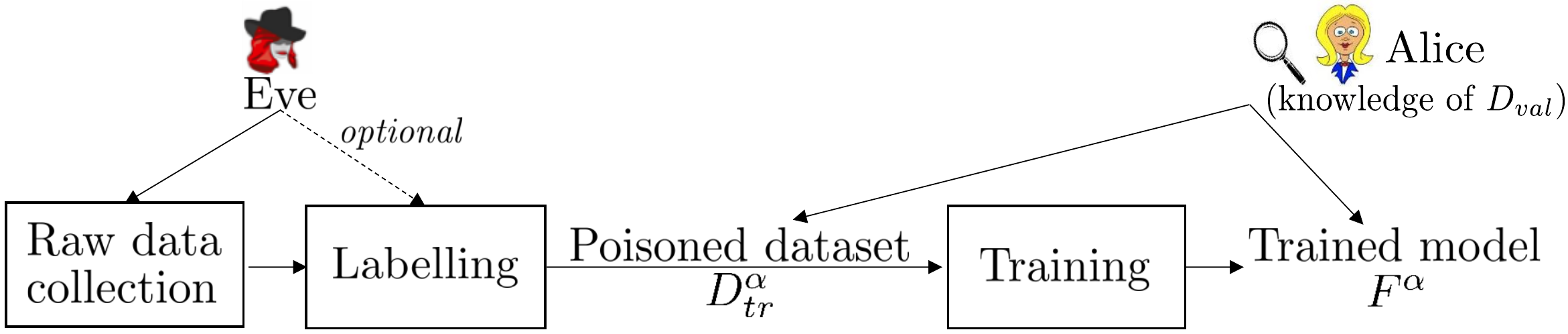}
	\caption{Threat model}
	\label{fig:t_model}
	\vspace{-2mm}
\end{figure}

The threat model considered in this work is illustrated in Fig. \ref{fig:t_model}. The attacker Eve interferes with
the data collection process, by poisoning a fraction $\alpha$ of the training dataset, possibly modifying the labels of the poisoned samples.
Alice, plays the role of the trainer, defining the model architecture, learning algorithm, and hyperparameters, and training the model.
Alice also plays the role of the defender: she inspects the training dataset and the deployed model to detect the possible presence of poisoned samples in the training set. The exact goal, knowledge and capabilities of the defender are detailed in the following.
%

\textbf{Defender's goal}: Alice aims at revealing the presence of poisoned samples in the training dataset $D^{\alpha}_{tr}$, if any, and identify them\footnote{For sake of simplicity, we use the notation $D^{\alpha}_{tr}$ for the training set under inspection, even if, prior to inspection, we do not know if the set is poisoned or not. For benign dataset we simply have $\alpha = 0$.}.
Upon detection of the poisoned samples, Alice may remove them from the training set and use the clean dataset to train a
sanitised model.

Formally, the core of the CCA-UD defence consists of a detector, call it $det()$,
defined as follows. For every subset $D_{tr,i}^{\alpha}$ of the training dataset $D_{tr}^{\alpha}$, $det (D_{tr,i}^{\alpha})=(P_i, B_i),$
%
%
where $P_i$ and $B_i$ are the sets with the samples in class $i$ judged  by $det()$ to be, respectively, poisoned and benign.
Extending the above functionality to all the classes in the input domain of the classifier,
we have $det (D_{tr}^{\alpha})= \{(P_i,B_i), i = 1,...,l\}$.
%
%
Clearly, for a non-poisoned dataset, \CH{ideally, we should have $P_i = \emptyset~~\forall i$, since no sample is poisoned in any class}.

\textbf{Defender's knowledge and capability}: Alice can inspect the training dataset $D_{tr}^{\alpha}$, and has white-box access to the trained model $F^{\alpha}$. Moreover, Alice has a small benign validation dataset $D_{val}$, with a small number of non-poisoned samples of every class. This is a requirement common to other methods like SCAn, Beatrix (while AC and CI do not require it).

\subsection{The proposed CCA-UD defence}
\label{sec.algorithm}



\begin{figure*}
	\centering
	\includegraphics[width=2\columnwidth]{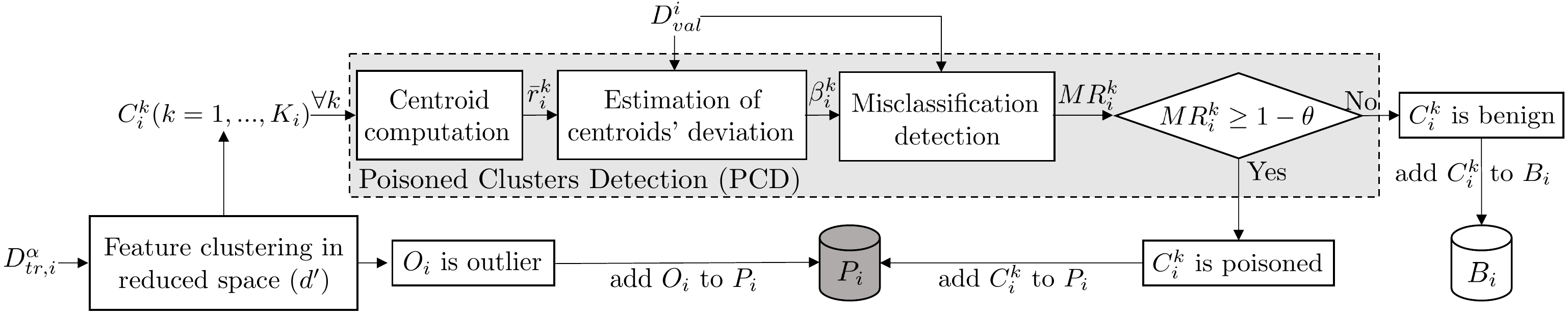}
	\caption{Workflow of the CCA-UD defence}
	\label{fig:workflow}
	\vspace{-4mm}
\end{figure*}

CCA-UD consists of two main blocks: \textit{feature clustering} and \textit{Poisoned Cluster Detection (PCD)}, shown in Fig.~\ref{fig:workflow} and detailed in Sections \ref{sec:cluster} and \ref{sec:TRA}\footnote{The code implementing CCA-UD is available at the address \url{https://github.com/guowei-cn/CCA_UD-universal-training-level-defence.git}.}.

Feature clustering relies on the DBSCAN algorithm \cite{EsterKSX96}.
DBSCAN splits a set of points into $K$ clusters and possibly few outliers, where $K$ is automatically determined by counting the areas with high sample density.
%
We refer to \cite{EsterKSX96} for  more information on the DBSCAN method.
The performance of DBSCAN is affected by the choice of the parameters involved in its definition, that is $minPts$ (the threshold on the number of neighbours used to define core points, which determines the dense regions constituting the clusters) and $\epsilon$ (distance defining the neighbourhood), whose setting depends on the problem at hand. The influence of such parameters on CCA-UD and the way we set them are described in Section \ref{sec:threshold}.
We decided to use density-based clustering since it works well also in the presence of clusters with unbalanced size \cite{rokach2005clustering}, and because it provides an automatic way to determine the number of clusters\footnote{DBSCAN is one of the most popular density-based clustering algorithms, other choices, like OPTICS \cite{ankerst1999optics} and HDBSCAN \cite{campello2015hierarchical}) would work as well.}.



\subsubsection{Dimensionality reduction and feature clustering}
\label{sec:cluster}

Sample clustering works in three steps.
In the first step, for every class $i$, we compute the feature representations of all the samples in $D_{tr,i}^{\alpha}$, namely $\{f_{1}^{\alpha}(x_j), x_j\in D_{tr,i}^{\alpha}\}$. Vector $f_{1}^{\alpha}(x_j)$ is a $d$-dim vector.
Secondly, we reduce the dimension of the feature space  from $d$ to $d'$ via Uniform Manifold Approximation and Projection (UMAP)~\cite{mcinnes2018umap}.
Finally, we apply DBSCAN to split $D_{tr,i}^{\alpha}$ into multiple clusters $C_{i}^{k}(k=1,...,K_i)$.
In addition to clusters,  DBSCAN (may) also return a number of outliers.
The set with the outlier samples, referred to as $O_i$, is directly added to $P_i$. The outlier ratio for the class $i$ is denoted by $\zeta_i=\frac{|O_i|}{|D_{tr,i}^{\alpha}|}$. With the hyperparameters ($d'$, $minPts$ and $\epsilon$) we have chosen, $\zeta_i$ is usually very small
(see Table S1 reported in the supplementary material).

Notably, in \cite{abs-2302-12758}, the separability of poisoned and benign features (after PCA reduction) in different layers is investigated, to
understand if some layers are more effective than others for the discrimination, and a method is proposed to find the layer where  poisoned samples and benign samples are most distinguishable. According to our experiments, however, when UMAP~\cite{mcinnes2018umap} algorithm  is used for dimensionality reduction, poisoned samples can be distinguished well from benign samples in all layers, and CCA-UD achieves similar performance regardless of the layer where the analysis is carried out. A visualisation of feature separability
after PCA and UMAP is shown in the supplementary material (Section II). \BTcomm{@David: 1) check if the above rewriting is ok. 2) what does it meas that 'the separability at the various layers is similar'? Is it really the case that in every layer the feature maps are separated in the same way? so the layer we are using (after conv part) is not better than another layer ....If this is the case we could choose a layer with lower dimensionality and do everything there that would be less complexy.....Given this observation, we should perhaps try to motivate our choice to work in the  layer after conv.} 
\CH{We observe that the dimensionality reduction is exploited only to run DBSCAN, while in the other steps the full feature space is considered.
Reducing the dimensionality of the features before applying DBSCAN  permits to reduce the time complexity of the algorithm and at the same time avoids curse of dimensionality problems \cite{koppen2000curse}, occurring when clustering is applied in high-dimensional spaces  (see the experiments reported in the supplementary material -  Section {IV} - supporting this claim).  }



\subsubsection{Poisoned cluster detection (PCD)}
\label{sec:TRA}

To determine if a cluster $C_{i}^{k}$ is poisoned or not, we first compute an average representation of the samples in $C_{i}^{k}$, i.e., the cluster's centroid. Then, we check whether the centroid contains a feature component that causes a misclassification in favour of class $i$ when added to the features of benign samples of the other classes.
More specifically,
%
we first calculate the centroid of $C_i^k$ as $\bar{r}_{i}^{k}=E[f_{1}^{\alpha}(x_j) |x_j\in C_{i}^{k}]$, where $E[\cdot]$ denotes component-wise sample averaging. Vector $\bar{r}_{i}^{k}$ is a $d$-dimensional vector\footnote{We remind that, although clustering is applied in the reduced-dimension space, the analysis of the clusters is performed in the full features space.}.
Then, we compute the deviation of $\bar{r}_{i}^{k}$ from the centroid of class $i$ computed on a set of benign samples:
\begin{equation}
	\label{eq:recoverTR}
	\beta_{i}^{k}= \bar{r}_{i}^{k}-E[f_{1}^{\alpha}(x_j)|{x_j\in D_{val}^{i}}],
\end{equation}
where $D_{val}^{i}$ is the $i$-th class of the benign set $D_{val}$.

Finally, we check if $\beta_{i}^{k}$ causes a misclassification error in favour of class $i$
when it is added to the feature representation of the benign samples in $D_{val}$ belonging to any class but the $i$-th one. The corresponding misclassification ratio is computed as follows:
%
\begin{equation}
\label{eq:MR}
	MR_i^k=\dfrac{\sum_{x_j\in D_{val}/D_{val}^{i}}\mathds{1}\big\{f_{2}^{\alpha}\Big(\delta(f_{1}^{\alpha}(x_j)+\beta_{i}^{k})\Big)\equiv i\big\}}{|D_{val}/D_{val}^{i}|},
\end{equation}
where $\mathds{1}\{\cdot\}$ is the indicator function (outputting 1 when the condition is satisfied and zero otherwise),
$D_{val}/D_{val}^{i}$ represents the validation dataset excluding the samples from class $i$, and $\delta$ is a ReLu operator that ensures that $f_{1}^{\alpha}(x_j)+\beta_{i}^{k}$ is a correct vector in the latent space (see Section \ref{sec:notation}).

For a given threshold $\theta$, if $MR_{i}^{k}\geq 1-\theta$ \footnote{We defined the threshold as $1-\theta$ to ensure that $TPR$ and $FPR$ increase with the growth of $\theta$ as for AC and CI, so to ease the comparison between the various defences.}, the corresponding $C_{i}^{k}$ is judged to be poisoned and its elements are added to $P_i$. Otherwise, the cluster is considered benign and its elements are added to $B_i$. Given that  $MR_{i}^{k}$ takes values in $[0,1]$, the threshold $\theta$ is also chosen in this range.

\subsubsection{Expected behaviour of CCA-UD for clean- and corrupted-label attacks}

An intuition of the idea behind CCA-UD and the reason why the detection of poisoned samples works for both corrupted- and clean-label attacks is given in the following. Let us focus first on the clean-label attack scenario. If cluster $C_{i}^{k}$ is poisoned, the centroid $\bar{r}_{i}^{k}$ contains the features of the trigger in addition to the feature of class $i$.
Then, arguably, the deviation of the centroid from the average representation of class $i$ is a significant one. Ideally,
subtracting to $\bar{r}_{i}^{k}$ the average feature representation of the $i$-th class, obtaining $\beta_{i}^{k}$, isolates the trigger features.
The basic idea behind CCA-UD is that the trigger features in $\beta_{i}^{k}$ will cause a misclassification in favour of class $i$ when added to the features of benign samples of the other classes.
%
On the contrary, if cluster $C_{i}^{k}$ is benign, the centroid $\bar{r}_{i}^{k}$ approximates the average feature representation of the $i$-th class
and then $\beta_{i}^{k}$ has a very small magnitude. In this case, $\beta_{i}^{k}$ accounts for normal intra-class fluctuation of the features and its addition to benign samples is not expected to induce a misclassification.

Similar arguments, with some noticeable differences, hold in the case of corrupted-label attacks. As before, for a benign cluster $C_{i}^{k}$, the centroid $\bar{r}_{i}^{k}$  approximates the average feature representation of the $i$-th class and then $\beta_{i}^{k}$ corresponds to minor intra-class variations. In the case of a poisoned cluster $C_{i}^{k}$, the cluster now includes mislabeled samples of the other classes (different from $i$) containing the triggering signal. In this way, the cluster representative contains features of the original classes in addition to the features of the triggering signal.
%
Note that even if the clustering algorithm splits the poisoned samples across more than one cluster, the deviation vector $\beta_{i}^{k}$ of poisoned clusters will contain the features of the triggering signal (possibly in addition to features accounting for {\em difference} between the original class $i$ and the target class $t$). Given that the network has been trained to {\em recognise} the triggering signal as a distinguishing feature of class $t$,  once again, the addition of the deviation vector to benign samples is likely to cause a misclassification in favour of class $t$.

\begin{figure*}
\begin{subfigure}{0.48\textwidth}
\centering
   \includegraphics[width=0.95\linewidth]{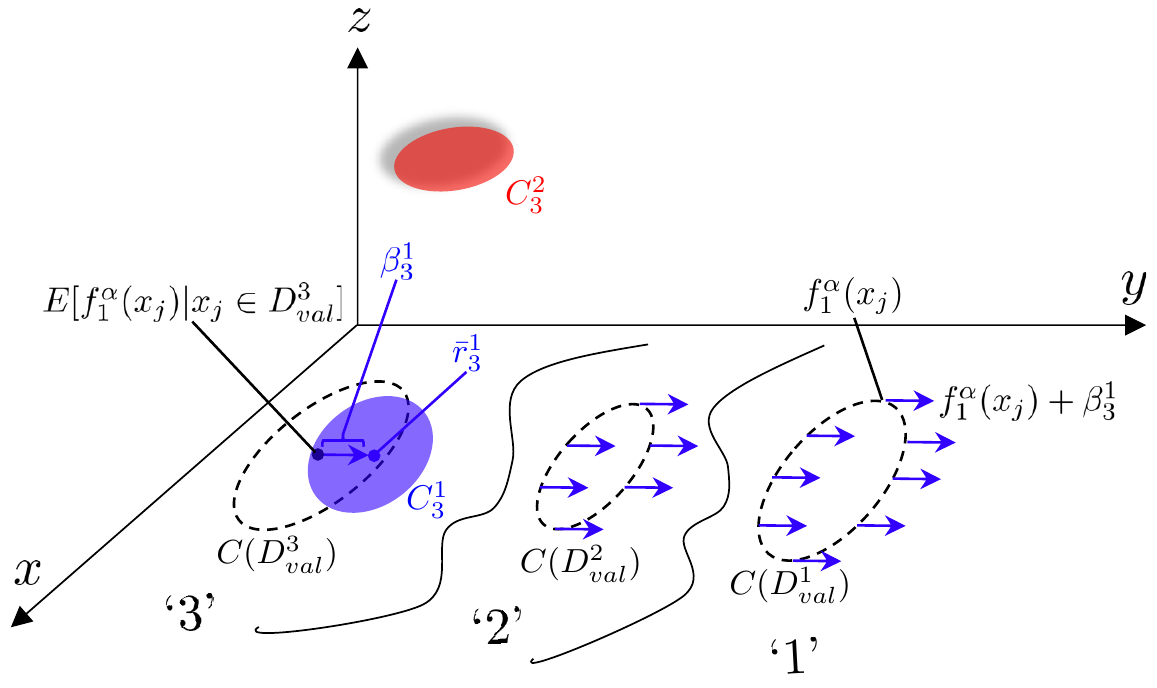}
   \label{fig:illustration1}
\end{subfigure}
\begin{subfigure}{0.518\textwidth}
\centering
   \includegraphics[width=0.95\linewidth]{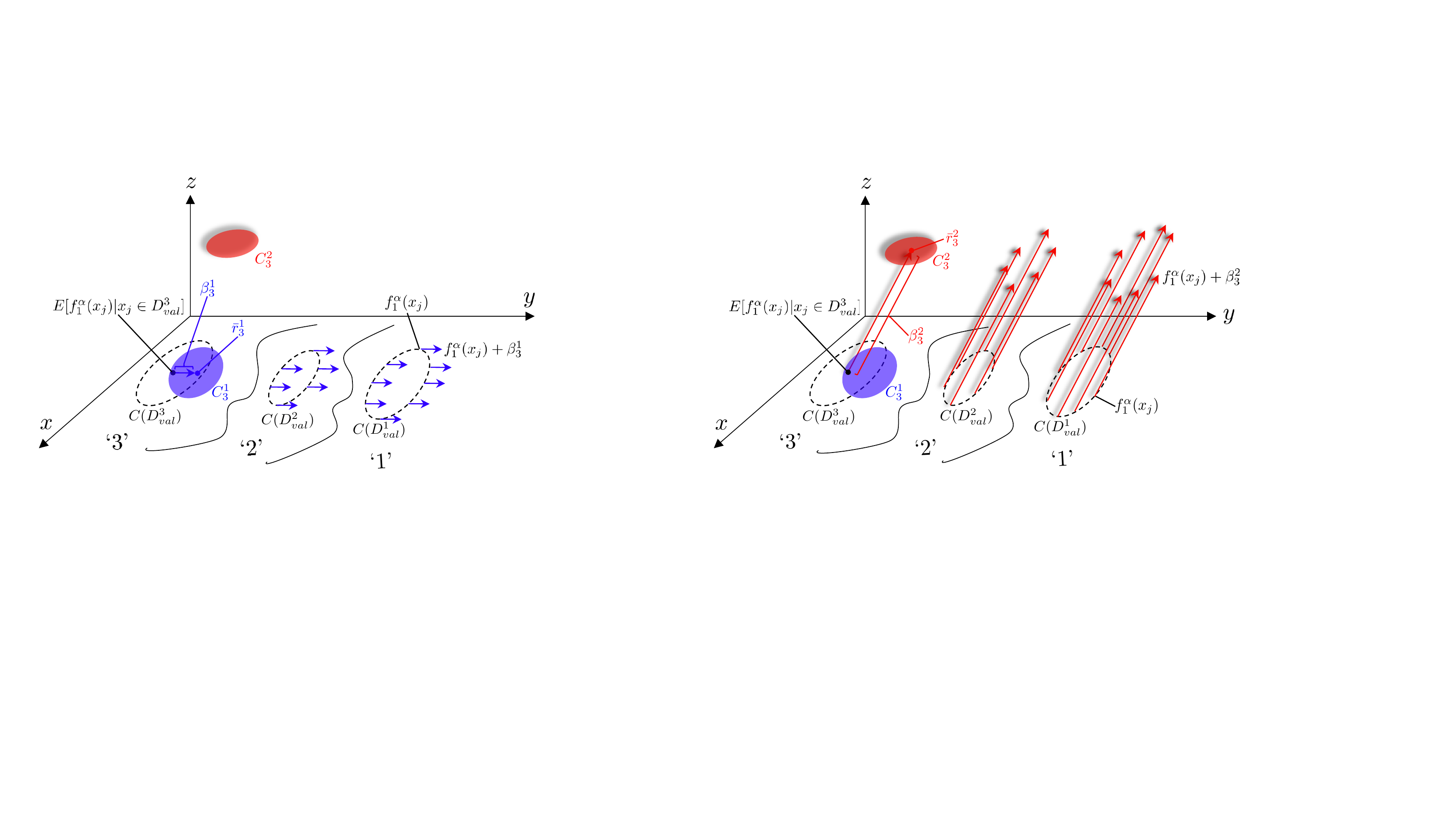}
   \label{fig:illustration2}
\end{subfigure}
\caption{Simplified illustration of PCD (clean-label case). For class `3' as the poisoned class, DBSCAN identifies two clusters: $C_3^1$ (benign) and $C_3^2$(poisoned).
When PCD is applied to $C_3^1$ (left part), the deviation from the set of benign samples of class $i$ ($C(D^3_{val})$) has a small amplitude and lies on $x$-$y$ plane, 
which cannot cause misclassification when added to the benign samples in $D^1_{val}$ and $D^2_{val}$. Instead, the deviation vector of $C_3^2$ (right part), containing a significant component in the $z$ direction, can cause misclassification.}
   \label{fig:pictorial}
   \vspace{-4mm}
\end{figure*}

The situation is pictorially illustrated in Fig. \ref{fig:pictorial} for a 3 dimension case, in the case of a clean-label attack (a similar picture can be drawn in the corrupted label case).
Class `3' corresponds to the poisoned class. Due to the presence of the backdoor, the poisoned samples are characterised by a non-null feature component along the $z$ direction. Due to the presence of such a component, the backdoored network classifies those samples in class `3'. On the contrary, benign samples lie in the $x$-$y$ plane.
When CCA-UD is applied to the samples labelled as Class `3', DBSCAN identifies two clusters, namely $C_3^1$ and $C_3^2$, where the former is a benign
cluster and the latter is a poisoned cluster containing a non-null $z$-component. When PCD module is applied to $C_3^1$ (left part in the figure), the deviation from the set of benign samples of class $i$ ($\beta_3^1$), has a small amplitude and lies in the $x$-$y$ plane, hence when $\beta_3^1$ is added to the other clusters it does not cause a misclassification error. Instead, when the PCD module is applied to $C_3^2$ (right part in the figure), the deviation vector ($\beta_3^2$) contains a significant component in the $z$ direction, causing a misclassification when added to the benign samples in $D^1_{val}$ and $D^2_{val}$.

It is worth stressing that CCA-UD indirectly exploits a general behaviour induced by backdoor attacks, that also works for sample-specific attacks when the samples are poisoned by applying a specific processing (e.g. a deformation/warping \cite{Duchon76}).
In this case, the backdoored network associates the traces of the specific processing applied to the input samples to one or more peculiar features (fingerprint), and uses such an evidence to misclassify these samples as belonging to the target class (it is possible that, in this case, the features of the poisoned samples are characterised by a larger variability with respect to the case in which the attack relies on a fixed triggering signal).

\subsubsection{Discussion}

We observe that the universality of CCA-UD essentially derives from the use of DBSCAN and from the generality of the proposed strategy for PCD.
Firstly, in contrast to $K$-means, DBSCAN can handle unbalanced clusters, ensuring good performance also when the poisoning ratio $\alpha$ is small or when the number of poisoned samples is larger than the number of benign samples.
Secondly, CCA-UD also works when the class samples have large intra-variability, since the ultimate decision on the presence of a cluster with poisoned samples is made is made by observing the {\em corrupting capabilities} of the cluster samples.
In addition, in the presence of large intra-class variability, DBSCAN groups the data of a benign class into multiple clusters (a large $K_i$, $K_i > 2$, is estimated by DBSCAN), that are then detected as benign clusters.
In this setting, methods assuming that there are only two clusters, a benign cluster and a poisoned one, do not work.

Finally, we observe that thanks to the fact that $K_i$ is directly estimated by DBSCAN,
in principle our method can also work in the presence of multiple triggering signals \cite{salem2022dynamic,xue2020one}.
In this case, the samples poisoned by different triggers would cluster in separate clusters, that would all be detected as poisoned by CCA-UD. 

\section{Experimental methodology}
\label{sec:exp}




\subsection{Evaluation metrics}
\label{sec:metric}

The performance of the backdoor attacks is evaluated by providing the accuracy of the backdoored model $F^{\alpha}$ on benign data and the success rate of the attack when the model is tested on poisoned data. The two metrics we use to do that are:
\begin{itemize}
	\item
Accuracy ($ACC$), measuring the probability of a correct classification of benign samples, and calculated as follows:
	\begin{equation}
 \text{$ACC$}=  \sum_{i = 1}^l\sum_{x_j\in D_{ts,i}}\mathds{1}\{F^{\alpha}(x_j)\equiv i\}/|D_{ts}|;
	\end{equation}

	\item Attack Success Rate ($ASR$), measuring the probability that the triggering  signal $\upsilon$  activates the desired behaviour of the backdoored model $F^{\alpha}$, computed as follows:
	\begin{equation}
		ASR=\frac{1} {|D_{ts}/D_{ts,t}|}\sum_{x_j\in D_{ts}/D_{ts,t}}\mathds{1}\{F^{\alpha}(\mathcal{P}(x_{j}, \upsilon))\equiv t\},
	\end{equation}
 where $D_{ts}/D_{ts,t}$ is the test dataset excluding the samples from class $t$.
\end{itemize}
In our experiments, a backdoor attack is considered successful when $ASR$ is greater than 0.90 and the $ACC$ of the poisoned model is similar to that of a model trained over benign samples (in our experiments such a difference is smaller than 0.01).
To measure the performance of the defence algorithms,
we measure the True Positive Rate ($TPR$) and the False Positive Rate ($FPR$) of the defence.
Actually, when $i$ corresponds to a benign class, there are no poisoned samples in $D_{tr,i}^{\alpha}$  and  only the $FPR$ is computed.
More formally, let $GP_i$ (res. $GB_i$) define the set of ground-truth poisoned (res. benign) samples in $D_{tr,i}^{\alpha}$. We define the $TPR$ and $FPR$ on $D_{tr,i}^{\alpha}$ as follows:
\begin{equation}
TPR(D_{tr,i}^{\alpha})=\frac{|P_i \cap GP_i|}{|GP_i|}, FPR(D_{tr,i}^{\alpha})=1-\frac{|B_i \cap GB_i|}{|GB_i|}.
\label{eq:TPR_FPR}
\end{equation}
Given that benign classes may exist for both poisoned and benign datasets\footnote{The backdoor attack does not need to target all classes in the input domain.}, we need to distinguish between these two cases. Hence, we introduce the following definitions:
\begin{itemize}
	\item Benign Class of Benign dataset ($BC_B$):
 a class of a clean dataset. In this case $\alpha = 0$ and $D_{tr,i}^{\alpha}$ includes only benign samples.
	\item Benign Class of Poisoned dataset ($BC_P$):
 a  benign class of a poisoned dataset, that is, a class in a poisoned dataset different from the target class. Also in this case, $D_{tr,i}^{\alpha}$ includes only benign samples.
\end{itemize}
The difference between $BC_B$ and $BC_P$ is that in the former case $F^{\alpha}$ is a clean model, while in the latter it is backdoored.
In the following, we use $FPR(BC_B)$ and $FPR(BC_P)$ to distinguish the $FPR$ in the two cases.

Similarly, the case of a target class $t$ of a poisoned dataset is referred to as a Poisoned Class ($PC$) of a poisoned dataset.
In this case, $D_{tr,i=t}^{\alpha}$ includes both poisoned and benign samples,
then we compute and report $TPR(PC)$ and $FPR(PC)$.
$TPR$ and $FPR$ depend on the choice of the threshold $\theta$. Every choice of threshold defines a different operating point of the detector. In order to get a global view of the performance of the tested systems, then, we also provide the $AUC$ value, defined as the Area Under the Curve obtained by varying the value of the threshold and plotting $TPR$ as a function of $FPR$. 

According to the definitions in Eq. \eqref{eq:TPR_FPR}, the false positive and true positive rates are computed for each cluster. For the sake of simplicity, we will often report average values.
For the case of benign clusters of a benign dataset, the average value,
denoted by $\overline{FPR}(BC_B)$,
is calculated by averaging over all the classes of the benign training dataset.
To compute the average metrics in the case of $BC_P$ and $PC$,
we repeat the experiments several times
by poisoning different target classes with various poisoning ratios $\alpha$ in the range (0, 0.55] for every target class, and by using the poisoned datasets to train the backdoored models\footnote{Only successful backdoor attacks are considered to measure the performance in the various cases.}. 
%
Then, the average $\overline{FPR}(BC_P)$ is computed by averaging the performance achieved on non-target classes of all the poisoned training datasets.
For the $PC$ case, the average metrics $\overline{FPR}(PC)$, $\overline{TPR}(PC)$ and $\overline{AUC}$ are computed by averaging the values measured on the target classes of the poisoned training datasets.
We also measured the average performance achieved for a fixed poisoned ratio $\alpha$, by varying only the target class  $t$.
The notation $\overline{FPR}_{\alpha}(BC_P)$, $\overline{FPR}_{\alpha}(PC)$, $\overline{TPR}_{\alpha}(PC)$, $\overline{AUC}_{\alpha}$ is used in this case to highlight the dependence on $\alpha$.

\begin{figure*}
	\centering
	\includegraphics[width=2.03\columnwidth]{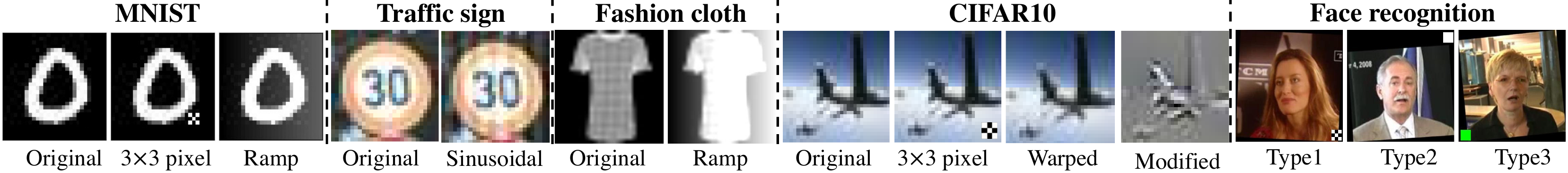}
	\caption{Triggering signals. On CIFAR10, the `Modified' image represents the difference between the original and warped image. }
	\label{fig:trigger}
	\vspace{-4mm}
\end{figure*}

\subsection{Network tasks and attacks}
\label{sec:settings}

To validate the effectiveness of CCA-UD, we carried out extensive experiments considering the following classification tasks and attacks.


\subsubsection{MNIST classification}
In this set of experiments, we trained a model to classify the digits in the MNIST dataset,
which includes $n=10$ digits (classes) with 6000 binary images per class. The size of the images is $28\times 28$.
The  architecture  used for the task is a 4-layer network \cite{4layer}.
The feature representation of dimensionality 128 is obtained from the input of the final fully-connected layer.

Regarding the attack setting, three different backdoor attacks have been considered, as detailed below. For each setting, the training dataset is poisoned by considering 16 poisoning ratios $\alpha$ chosen in $(0, 0.55]$. For each $\alpha$, 10 different poisoned training datasets are generated by choosing different classes as the target class.

\begin{itemize}
	 \item Corrupted-label attack, with a 3$\times$3 pixel trigger (\textit{3$\times$3 corrupted}): the backdoor is injected by adding a 3$\times$3 pixel pattern to the corrupted samples, as shown in Fig. \ref{fig:trigger}, and modifying the sample labels into that of the target class. 
    \item Corrupted-label attack, with ramp trigger (\textit{ramp corrupted}): Eve performs a corrupted-label backdoor attack using a horizontal ramp signal~\cite{BarniKT19} as a trigger (see Fig. \ref{fig:trigger}). The ramp signal is defined as $\upsilon(i,j)=j\Delta/W$, $1\leq i\leq H$, $1\leq j\leq W$, where $H\times W$ is the size of the image. The parameter $\Delta$ controls the slope (and strength) of the ramp trigger signal, \CH{with a larger value leading to a more visible trigger (and to a more effective attack).} We set $\Delta=40$ in the experiments\CH{, that represents a good tradeoff between visibility and effectiveness of the attack \cite{BarniKT19}}.
    \item Clean-label attack, with 3$\times$3 pixel trigger (\textit{3$\times$3 clean}): the 3$\times$3 pixel trigger signal is utilised  to perform a clean-label attack.
\end{itemize}

\subsubsection{Traffic signs}
For the traffic sign classification task,
we selected 16 different classes from the GTSRB dataset, namely, the most representative classes in the dataset, including 6 speed-limit, 3 prohibition, 3 danger, and 4 mandatory signs. Each class has 1200 colour images with size 28$\times$28. The model architecture used for training is based on ResNet18 
%
The feature representation is extracted from the 17-th layer, that is, before the FC layer, after an average pooling layer and ReLu activation. With regard to the attack, we considered the corrupted-label scenario. The triggering signal is a horizontal sinusoidal signal,
defined as $\upsilon(i,j)=\Delta \sin(2\pi jf/W)$, $1\leq i\leq H$, $1\leq j \leq W$, where $H\times W$ is the size of input image.
The parameters $\Delta$ and $f$ are \CH{the amplitude and frequency of the sinusoidal function, which control the strength and frequency of the sinusoidal signal.}
In our experiment, we set $\Delta=20$ and $f=6$. The sinusoidal trigger is shown in Fig. \ref{fig:trigger}.
As before, for a given $\alpha$, the network is trained on 16 poisoned datasets, each time considering different target classes.

\subsubsection{Fashion clothes}
The Fashion-clothes dataset includes 10 classes of grey-level cloth images, each class consisting of 6000 images of size {28$\times$28}. 
The model architecture used for the classification is based on AlexNet {with input size equal to 224$\times$224}. 
%
The representation used by the backdoor detector is extracted from the 5-th layer, at the output of the ReLu activation layer before the first FC layer.
With regard to the attack, the poisoned samples are generated by performing the attack in a clean-label setting. A ramp trigger with $\Delta=256$ is used to implement the attack, as shown in Fig. \ref{fig:trigger}.
Once again, for each choice of $\alpha$, the backdoor attack is repeated 10 times, each time considering a different target class.



Furthermore, to prove the universality of the proposed method, we also run some experiments considering more complex tasks and realistic datasets, namely  CIFAR10 classification and face recognition, and with different backdoor attacks relying on source-specific and sample-specific triggers. The setting used and the results achieved in these cases are reported in Section \ref{sec.others}.

For all the classification tasks, the benign validation dataset $D_{val}$ is obtained by randomly selecting 100 samples from all the classes in the dataset.

\subsection{Parameters setting and state-of-the-art comparison}
\label{sec:threshold}
To implement CCA-UD, we have to set the following parameters: the reduced dimension $d'$ for the clustering, the parameters of the DBSCAN algorithm, namely $minPts$ and $\epsilon$, and finally the threshold $\theta$ used by the clustering poisoning detection module.
In our experiments, we set  $d' = 2$,  $minPts = 20$ and $\epsilon = 0.8$. This is the setting that, according to our experiments, achieves the best performance with the minimum complexity for the clustering algorithm (being $d'=2$)\footnote{\CH{Notably, the same setting works for all the cases, namely for all the tasks, architectures and attacks we have considered in our experiments.}}. The effect of these parameters on the result of clustering and detection performance is evaluated by the ablation study reported in the supplementary material (Section~I).

We compared our method with the state-of-the-art methods mentioned in Section \ref{sec:related}, namely, AC~\cite{ChenCBLELMS19}, CI~\cite{Xiang0K19}, SCAn~\cite{Tang0TZ21} and Beatrix~\cite{MaWSXWX23}.

With regard to  $\theta$, as mentioned before, AC, CI, SCAn, Beatrix and CCA-UD involve the setting of a threshold for poisoning detection.
Specifically, in AC the relative size of the class ratio is thresholded, while in CI we vary the prediction change rate between filtered and non-filtered samples. For SCAn, the threshold applies to the hypothesis testing statistics used to judge if a class is poisoned or not (the smallest size cluster obtained after the application of the LDA represents the set of poisoned data). Finally, for Beatrix, we varied the anomaly detection threshold
\BTcomm{If we will run out of space we can avoid giving the details of what is the threshold in the various cases (I like having it however many papers do not report these info)}

For a fair comparison, we set the threshold in the same way for all the methods.
In particular, we set $\theta$ 
by fixing the average $FPR$ on the validation dataset (consisting of benign samples). For each class $D_{val}^i$, we calculate the $FPR(D_{val}^i)$ value, and its average counterpart is $\overline{FPR}(D_{val})=\sum_i FPR(D_{val}^i)/l$. We chose the threshold in such a way to minimise the distance from the target rate.
Formally, by setting the target false positive rate to 0.05, the threshold $\theta^*$ is determined as:
\begin{equation}
\theta^*=\arg\min_{\theta}\big|0.05-\overline{FPR}(D_{val})\big|.
\label{eq:det_th}
\end{equation}

\CH{The parameters
of CCA-UD and their settings are summarized in  Table \ref{tab:hp}.}

\begin{table}[h]
\caption{\CH{Setting of DBSCAN parameters in CCA-UD.}}
\label{tab:hp}
\centering
 \begin{tabular}{|p{1cm}|p{1.4cm}|p{5cm}|}
  \hline
Parameter & Setting & Meaning  \\ \hline  \hline
 $d'$ & 2 & Dimension of the reduced space (UMAP)  \\ \hline
 $\epsilon$ & 0.8 & Neighbourhood radius (DBSCAN) \\ \hline
 $\theta^*$ & Set via \eqref{eq:det_th} & Threshold used in the PCD module \\ \hline
 $minPts$ & 20 & Minimal sample numbers in the neighbourhood of a core point (DBSCAN) \\ \hline
 \end{tabular}
 \vspace{-4mm}
\end{table}

\section{Experimental results}
\label{sec:results}

\subsection{Threshold setting}
\label{sec.thresholdresults}


Our experiments reveal that, for AC and CI, the threshold determined via Eq. \eqref{eq:det_th} 
does not lead to a good operating point when used on the test dataset (see  Table \ref{tab:mnist} for the  MNIST case), and a threshold that works well in all cases can not be found for AC and CI.
In particular,  while for SCAn, Beatrix and CCA-UD, the threshold $\theta^*$ set on the validation dataset yields a similar  $\overline{FPR}$ (around 0.05) in the $BC_B$, $BC_P$ and $PC$ cases, this is not true for AC and CI, for which  $\overline{FPR}(BC_B)$, $\overline{FPR}(BC_P)$ and $\overline{FPR}(PC)$ are often smaller than 0.05, reaching 0 in many cases. This leads to a poor $\overline{TPR}(PC)$.
In particular, with AC, when $\alpha > \theta^*$, both clusters are classified as benign, and then $\overline{TPR}_{\alpha}(PC)$ = $\overline{FPR}_{\alpha}(PC)=0$, even when the method would, in principle, be able to provide perfect discrimination ($\overline{AUC}_{\alpha}\approx 1$).
The difficulty in setting the threshold for AC and CI is also evident from the plots in Fig.~\ref{fig:exp_mnist},  that report the $\overline{FPR}$ and $\overline{TPR}$ values (that are averaged also on $\alpha$), for different values of the threshold $\theta$. From these plots,  we clearly see that a  threshold that works for all $\alpha$ can not be found for AC and CI.

Due to the difficulties encountered to set the detection threshold for AC and CI\footnote{Note that the problem of threshold setting is not addressed in the original papers, since different thresholds are used in the various cases.},
%
the results at $\theta^*$ for these methods are
not reported in the other cases,
for which we report only the $\overline{AUC}_{\alpha}$.
Note that the possibility to set a unique threshold on a benign dataset that also works on poisoned datasets is very important for the practical applicability of a defence.


\subsection{Results on MNIST}
\label{sec:resultM}


%

\begin{figure*}[h!]
\centering
\begin{subfigure}[b]{1\textwidth}
\centering
   \includegraphics[width=0.9\linewidth]{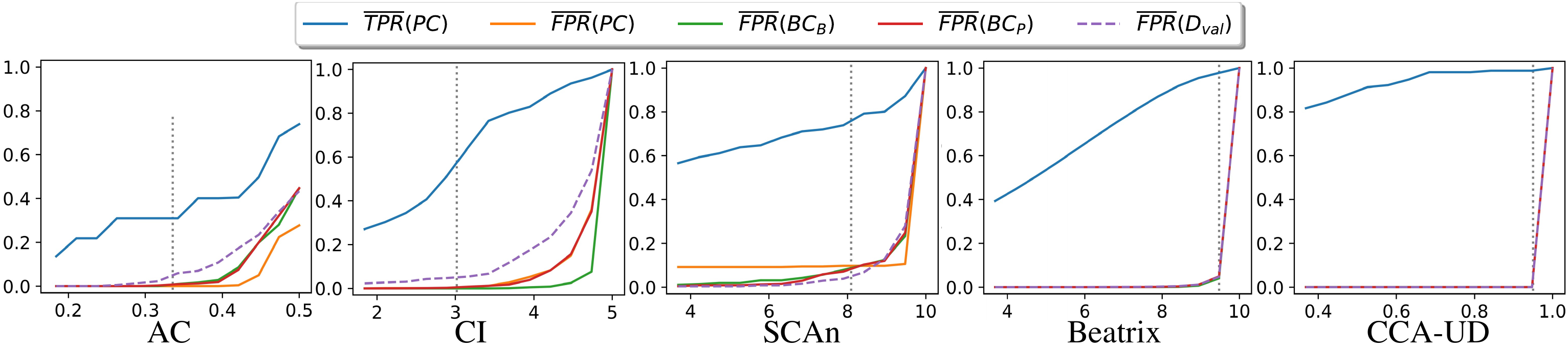}
   \caption{3$\times$3 corrupted attack}
\end{subfigure} \\
\begin{subfigure}[b]{1\textwidth}
\centering
   \includegraphics[width=0.9\linewidth]{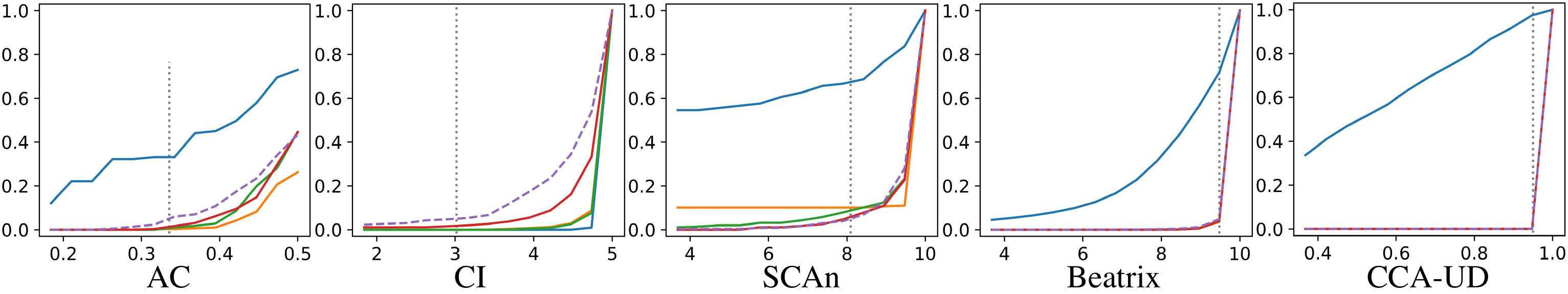}
   \caption{Ramp corrupted attack}
\end{subfigure} \\
\begin{subfigure}[b]{1\textwidth}
\centering
   \includegraphics[width=0.9\linewidth]{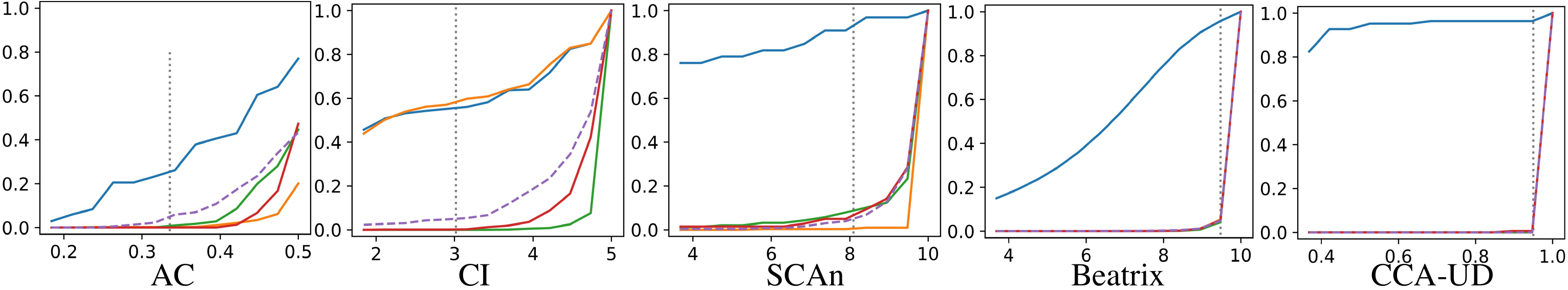}
   \caption{3$\times$3 clean attack}
\end{subfigure}
\caption{Average  performance of  AC and CI, SCAn, Beatrix, and CCA-UD for different the threshold against the three types of backdoor attacks implemented in the case of MNIST classification. 
The position of $\theta^*$ is indicated by a vertical dotted line. In each figure, the x-axis represents the threshold value, while the y-axis reports the values of $\overline{FPR}(BC_B)$, $\overline{FPR}(BC_P)$, $\overline{TPR}(PC)$ and $\overline{FPR}(PC)$ and $\overline{FPR}(D_{val})$}.
\label{fig:exp_mnist}
 \vspace{-6mm}
\end{figure*}

Performance is evaluated against the three types of backdoor attacks,
namely, \textit{3$\times$3 corrupted}, \textit{ramp corrupted}, and \textit{3$\times$3 clean}.
In Fig.~\ref{fig:exp_mnist}, the figures report the average performance of AC, CI, {SCAn, Beatrix,} and CCA-UD in the three cases.
The values of $\overline{FPR}(BC_B)$, $\overline{FPR}(BC_P)$, $\overline{TPR}(PC)$ and $\overline{FPR}(PC)$ are reported for each method, as a function of the detection threshold $\theta$.
The behaviour of $\overline{FPR}(D_{val})$, which is utilised to determine the threshold $\theta^*$ (at 0.05 of $\overline{FPR}(D_{val})$), is also reported.
The position of $\theta^*$ is indicated by a vertical dotted line\footnote{We verified that the threshold on the false positive rate set
on the validation dataset, also works on the test dataset, where we get the target $FPR$ (0.05 in our case).}.

By observing the figure, we see that CCA-UD outperforms the other methods in all the settings.
In the first setting, CCA-UD achieves $\overline{TPR}(PC)$ and $\overline{FPR}(PC)$ equal to {0.983} and 0.051 at the optimal threshold $\theta^{*}$, with $\overline{FPR}(BC_B)=0.051$ and $\overline{FPR}(BC_P)=0.050$. The performance of SCAn and Beatrix is a bit worse than CCA-UD, while the performance of AC and CI at their optimal threshold is very poor.
Similar results are achieved for the second and third settings.
In particular, for the second attack, CCA-UD achieves $\overline{TPR}(PC)$ and $\overline{FPR}(PC)$ equal to ({0.975}, 0.050) at $\theta^*$,
and (0.966,  {0.050})
for the third one.

\begin{table}
\caption{$\overline{AUC}$ of three methods with the various  attacks}
\label{tab:auc}
\centering
 \begin{tabular}{|c|c|c|c|}
 \hline
 \textbf{Method} & \textbf{3$\times$3 corrupted} & \textbf{Ramp corrupted} & \textbf{3$\times$3 clean} \\
 \hline
 AC & 0.728 & 0.733 & 0.785 \\ \hline
 CI &  0.964 & 0.178 & 0.488 \\ \hline
 SCAn & 0.848 & 0.868 & 0.984 \\ \hline
 Beatrix & 0.991 & 0.908 & 0.990 \\ \hline
 CCA-UD & {\bf 0.994} & {\bf 0.996} & {\bf 0.981} \\ \hline
 \end{tabular}
 \vspace{-4mm}
\end{table}

For a poisoned dataset, the $\overline{AUC}$ values obtained in the three settings are provided in Table \ref{tab:auc}.
From these results, we see that CI has good discriminating capability (with an AUC only slightly lower than CCA-UD) against the first attack, but fails to defend against the other two.
This is an expected behaviour since CI does not work when the triggering signal is robust against the average filter, as the ramp signal considered in the second attack, or clean-label attacks in the last setting.
SCAn and Beatrix instead have performance only slightly lower than our method.

\begin{table*}[h!]
 \caption{Performance
 for various poisoning ratios $\alpha$, against the three types of backdoor attacks for MNIST classification. $\overline{FPR}_{\alpha}(PC)$ and $\overline{TPR}_{\alpha}(PC)$ (indicated as $\overline{FPR}_{\alpha}$ and $\overline{TPR}_{\alpha}$ in the table) values are computed with $\theta = \theta^*$. In sub-table (c) cnt indicates the number of successful attacks in 10 repetitions.}
    \begin{subtable}[h]{1\textwidth}
    \centering
    \resizebox{1\columnwidth}{!}{
		\begin{tabular}{c|ccc|ccc|ccc|ccc|ccc|}
		\cline{2-16}
		&  \multicolumn{3}{c|}{\textbf{{AC}}} & \multicolumn{3}{c|}{\textbf{{CI}}} & \multicolumn{3}{c|}{\textbf{{SCAn}}}  & \multicolumn{3}{c|}{\textbf{{Beatrix}}} & \multicolumn{3}{c|}{\textbf{{CCA-UD}}} \\ \hline
		\multicolumn{1}{|c|}{\tiny{$\alpha$}} & \multicolumn{1}{c|}{\tiny $\overline{TPR}_{\alpha}$} & \multicolumn{1}{c|}{\tiny $\overline{FPR}_{\alpha}$} & \tiny{$\overline{AUC}_{\alpha}$} & \multicolumn{1}{c|}{\tiny $\overline{TPR}_{\alpha}$} & \multicolumn{1}{c|}{\tiny $\overline{FPR}_{\alpha}$} & \tiny{$\overline{AUC}_{\alpha}$} & \multicolumn{1}{c|}{\tiny $\overline{TPR}_{\alpha}$} & \multicolumn{1}{c|}{\tiny $\overline{FPR}_{\alpha}$} & \tiny{$\overline{AUC}_{\alpha}$} & \multicolumn{1}{c|}{\tiny $\overline{TPR}_{\alpha}$} & \multicolumn{1}{c|}{\tiny $\overline{FPR}_{\alpha}$} & \tiny{$\overline{AUC}_{\alpha}$} & \multicolumn{1}{c|}{\tiny $\overline{TPR}_{\alpha}$} & \multicolumn{1}{c|}{\tiny $\overline{FPR}_{\alpha}$} & \tiny{$\overline{AUC}_{\alpha}$} \\ \hline
		\multicolumn{1}{|c|}{0.025} & \multicolumn{1}{c|}{0.000} & \multicolumn{1}{c|}{0.000} & 0.563  & \multicolumn{1}{c|}{0.324}& \multicolumn{1}{c|}{0.022} & 0.876  & \multicolumn{1}{c|}{0.207} & \multicolumn{1}{c|}{0.061} & 0.810  & \multicolumn{1}{c|}{0.868} & \multicolumn{1}{c|}{0.057} &0.934 & \multicolumn{1}{c|}{ {0.908}}  & \multicolumn{1}{c|}{0.051} & 0.949  \\ \hline
		\multicolumn{1}{|c|}{0.050} & \multicolumn{1}{c|}{0.099} & \multicolumn{1}{c|}{0.000} &  0.628 & \multicolumn{1}{c|}{0.581} & \multicolumn{1}{c|}{0.001} & 0.977  & \multicolumn{1}{c|}{0.696} & \multicolumn{1}{c|}{0.000} & 0.987 & \multicolumn{1}{c|}{0.967} & \multicolumn{1}{c|}{0.048} & 0.992 & \multicolumn{1}{c|}{0.989} & \multicolumn{1}{c|}{0.050} & 0.994 \\ \hline
		\multicolumn{1}{|c|}{0.096} & \multicolumn{1}{c|}{0.395} & \multicolumn{1}{c|}{0.000} & 0.757&  \multicolumn{1}{c|}{0.654} & \multicolumn{1}{c|}{0.000} &0.996 &  \multicolumn{1}{c|}{0.992} & \multicolumn{1}{c|}{0.000} & 0.996  &  \multicolumn{1}{c|}{0.995} & \multicolumn{1}{c|}{0.046} & 0.999 &  \multicolumn{1}{c|}{0.999} & \multicolumn{1}{c|}{0.050} & 0.999 \\ \hline
		\multicolumn{1}{|c|}{0.134} &  \multicolumn{1}{c|}{0.792} & \multicolumn{1}{c|}{0.000} & 0.958&  \multicolumn{1}{c|}{0.559} & \multicolumn{1}{c|}{0.002} &0.990 &  \multicolumn{1}{c|}{0.994} & \multicolumn{1}{c|}{0.000} & 0.997 & \multicolumn{1}{c|}{0.998} & \multicolumn{1}{c|}{0.060} & 1.000 & \multicolumn{1}{c|}{0.999} & \multicolumn{1}{c|}{0.050} & 1.000 \\ \hline
		\multicolumn{1}{|c|}{0.186} & \multicolumn{1}{c|}{0.994}& \multicolumn{1}{c|}{0.000} &0.997&  \multicolumn{1}{c|}{0.577} & \multicolumn{1}{c|}{0.001} &0.985 & \multicolumn{1}{c|}{0.996} & \multicolumn{1}{c|}{0.000} &  0.998 &  \multicolumn{1}{c|}{0.999} & \multicolumn{1}{c|}{0.054} & 1.000 &  \multicolumn{1}{c|}{1.000} & \multicolumn{1}{c|}{0.050} &  1.000 \\ \hline
		\multicolumn{1}{|c|}{0.258} & \multicolumn{1}{c|}{0.993} & \multicolumn{1}{c|}{0.000} & 0.997 &  \multicolumn{1}{c|}{0.540} & \multicolumn{1}{c|}{0.070}  &0.961 & \multicolumn{1}{c|}{0.995} & \multicolumn{1}{c|}{0.000} & 0.998  &  \multicolumn{1}{c|}{0.999} & \multicolumn{1}{c|}{0.048} & 1.000 & \multicolumn{1}{c|}{1.000} &\multicolumn{1}{c|}{0.050} & 1.000 \\ \hline
		\multicolumn{1}{|c|}{0.359} & \multicolumn{1}{c|}{0.000} & \multicolumn{1}{c|}{0.000}  &0.998&  \multicolumn{1}{c|}{0.571} &\multicolumn{1}{c|}{0.005} &0.964 & \multicolumn{1}{c|}{0.996} & \multicolumn{1}{c|}{0.000} & 0.998 & \multicolumn{1}{c|}{0.999} & \multicolumn{1}{c|}{0.053} & 1.000 &  \multicolumn{1}{c|}{1.000} & \multicolumn{1}{c|}{0.050} &  1.000  \\ \hline
		\multicolumn{1}{|c|}{0.550} & \multicolumn{1}{c|}{0.000}  & \multicolumn{1}{c|}{0.000} & 0.001 & \multicolumn{1}{c|}{0.829} & \multicolumn{1}{c|}{0.000} &0.953 & \multicolumn{1}{c|}{0.002} & \multicolumn{1}{c|}{1.000} & 0.001  & \multicolumn{1}{c|}{0.996} & \multicolumn{1}{c|}{0.043} & 0.999 & \multicolumn{1}{c|}{1.000} & \multicolumn{1}{c|}{0.050} & 1.000 \\ \hline
	\end{tabular}}
	\caption{3$\times$3 corrupted}
	\label{tab:mnist_att1}
    \end{subtable}
    \begin{subtable}[h]{1\textwidth}
        \centering
        \resizebox{1\columnwidth}{!}{
        \begin{tabular}{c|ccc|ccc|ccc|ccc|ccc|}
		\cline{2-16}
		&  \multicolumn{3}{c|}{\textbf{{AC}}} & \multicolumn{3}{c|}{\textbf{{CI }}}  & \multicolumn{3}{c|}{\textbf{{SCAn}}} & \multicolumn{3}{c|}{\textbf{{Beatrix}}} & \multicolumn{3}{c|}{\textbf{{CCA-UD}}} \\ \hline
		\multicolumn{1}{|c|}{\tiny{$\alpha$}} & \multicolumn{1}{c|}{\tiny $\overline{TPR}_{\alpha}$} & \multicolumn{1}{c|}{\tiny $\overline{FPR}_{\alpha}$} & \tiny{$\overline{AUC}_{\alpha}$} & \multicolumn{1}{c|}{\tiny $\overline{TPR}_{\alpha}$} & \multicolumn{1}{c|}{\tiny $\overline{FPR}_{\alpha}$} & \tiny{$\overline{AUC}_{\alpha}$} & \multicolumn{1}{c|}{\tiny $\overline{TPR}_{\alpha}$} & \multicolumn{1}{c|}{\tiny $\overline{FPR}_{\alpha}$} & \tiny{$\overline{AUC}_{\alpha}$} & \multicolumn{1}{c|}{\tiny $\overline{TPR}_{\alpha}$} & \multicolumn{1}{c|}{\tiny $\overline{FPR}_{\alpha}$} & \tiny{$\overline{AUC}_{\alpha}$} & \multicolumn{1}{c|}{\tiny $\overline{TPR}_{\alpha}$} & \multicolumn{1}{c|}{\tiny $\overline{FPR}_{\alpha}$} & \tiny{$\overline{AUC}_{\alpha}$} \\ \hline
		\multicolumn{1}{|c|}{0.035} & \multicolumn{1}{c|}{0.050} & \multicolumn{1}{c|}{0.024} & 0.593 &  \multicolumn{1}{c|}{0.000} & \multicolumn{1}{c|}{0.008} &0.407 &  \multicolumn{1}{c|}{0.222} & \multicolumn{1}{c|}{0.020} & 0.948 &  \multicolumn{1}{c|}{0.423} & \multicolumn{1}{c|}{0.037} & 0.777 &  \multicolumn{1}{c|}{ {0.871}} & \multicolumn{1}{c|}{0.050} & 0.966  \\ \hline
		\multicolumn{1}{|c|}{0.050}& \multicolumn{1}{c|}{0.090} & \multicolumn{1}{c|}{0.028} & 0.593 & \multicolumn{1}{c|}{0.000} &  \multicolumn{1}{c|}{0.000} & 0.119 & \multicolumn{1}{c|}{0.300} & \multicolumn{1}{c|}{0.000} & 0.999 &  \multicolumn{1}{c|}{0.552} & \multicolumn{1}{c|}{0.036} & 0.851 &  \multicolumn{1}{c|}{0.914} & \multicolumn{1}{c|}{0.050} & 0.998 \\ \hline
		\multicolumn{1}{|c|}{0.096} & \multicolumn{1}{c|}{0.400}& \multicolumn{1}{c|}{0.000} &0.786&  \multicolumn{1}{c|}{0.000} &  \multicolumn{1}{c|}{0.000}  &0.216 & \multicolumn{1}{c|}{0.773} & \multicolumn{1}{c|}{0.000} & 0.999 &  \multicolumn{1}{c|}{0.739} & \multicolumn{1}{c|}{0.037} & 0.926 &  \multicolumn{1}{c|}{0.989} & \multicolumn{1}{c|}{0.050} & 0.998  \\ \hline
		\multicolumn{1}{|c|}{0.134} &  \multicolumn{1}{c|}{0.798}  & \multicolumn{1}{c|}{0.001} &0.962&   \multicolumn{1}{c|}{0.000} &  \multicolumn{1}{c|}{0.000} &0.142 & \multicolumn{1}{c|}{0.999} & \multicolumn{1}{c|}{0.000} & 0.999 &  \multicolumn{1}{c|}{0.732} & \multicolumn{1}{c|}{0.041} & 0.910 &  \multicolumn{1}{c|}{0.999} & \multicolumn{1}{c|}{0.050}&  {0.998} \\ \hline
		\multicolumn{1}{|c|}{0.186} & \multicolumn{1}{c|}{0.992} & \multicolumn{1}{c|}{0.003}  &0.995&  \multicolumn{1}{c|}{0.000} &  \multicolumn{1}{c|}{0.000}&0.179 & \multicolumn{1}{c|}{1.000} & \multicolumn{1}{c|}{0.000} & 1.000 &  \multicolumn{1}{c|}{0.781} & \multicolumn{1}{c|}{0.042} & 0.934 &  \multicolumn{1}{c|}{1.000} & \multicolumn{1}{c|}{0.050} &  1.000  \\ \hline
		\multicolumn{1}{|c|}{0.258} & \multicolumn{1}{c|}{0.999} & \multicolumn{1}{c|}{0.000} & 0.999 & \multicolumn{1}{c|}{0.000} &  \multicolumn{1}{c|}{0.000}&0.088 & \multicolumn{1}{c|}{1.000} & \multicolumn{1}{c|}{0.000} & 1.000 &  \multicolumn{1}{c|}{0.774} & \multicolumn{1}{c|}{0.035} & 0.941 & \multicolumn{1}{c|}{1.000} & \multicolumn{1}{c|}{0.050} & 1.000 \\ \hline
		\multicolumn{1}{|c|}{0.359} & \multicolumn{1}{c|}{0.000} & \multicolumn{1}{c|}{0.000}  &0.999& \multicolumn{1}{c|}{0.000} & \multicolumn{1}{c|}{0.000}  &0.144 & \multicolumn{1}{c|}{1.000} & \multicolumn{1}{c|}{0.000} & 1.000  &  \multicolumn{1}{c|}{0.834} & \multicolumn{1}{c|}{0.048} & 0.953 & \multicolumn{1}{c|}{1.000} & \multicolumn{1}{c|}{0.050} &  1.000 \\ \hline
		\multicolumn{1}{|c|}{0.550} & \multicolumn{1}{c|}{0.000} & \multicolumn{1}{c|}{0.000} & 0.002 &  \multicolumn{1}{c|}{0.000} &  \multicolumn{1}{c|}{0.000} &0.135 & \multicolumn{1}{c|}{0.006} & \multicolumn{1}{c|}{1.000} & 0.000 & \multicolumn{1}{c|}{0.873} & \multicolumn{1}{c|}{0.033} & 0.979 &  \multicolumn{1}{c|}{1.000} & \multicolumn{1}{c|}{0.050} & 1.000  \\ \hline
		\end{tabular}
		}
		\caption{Ramp corrupted}
    	\label{tab:mnist_att2}
     \end{subtable}
     \begin{subtable}[h]{1\textwidth}
     	\centering
     	\resizebox{1\columnwidth}{!}{
     	\begin{tabular}{cc|ccc|ccc|ccc|ccc|ccc|}
		\cline{3-17}
		&  & \multicolumn{3}{c|}{\textbf{{AC}}} & \multicolumn{3}{c|}{\textbf{{CI}}} & \multicolumn{3}{c|}{\textbf{{SCAn}}} & \multicolumn{3}{c|}{\textbf{{Beatrix}}} & \multicolumn{3}{c|}{\textbf{{CCA-UD}}} \\ \hline
		\multicolumn{1}{|c|}{\tiny{$\alpha$}} & cnt & \multicolumn{1}{c|}{\tiny $\overline{TPR}_{\alpha}$} & \multicolumn{1}{c|}{\tiny $\overline{FPR}_{\alpha}$} & \tiny{$\overline{AUC}_{\alpha}$} & \multicolumn{1}{c|}{\tiny $\overline{TPR}_{\alpha}$} & \multicolumn{1}{c|}{\tiny $\overline{FPR}_{\alpha}$} & \tiny{$\overline{AUC}_{\alpha}$} & \multicolumn{1}{c|}{\tiny $\overline{TPR}_{\alpha}$} & \multicolumn{1}{c|}{\tiny $\overline{FPR}_{\alpha}$} & \tiny{$\overline{AUC}_{\alpha}$} & \multicolumn{1}{c|}{\tiny $\overline{TPR}_{\alpha}$} & \multicolumn{1}{c|}{\tiny $\overline{FPR}_{\alpha}$} & \tiny{$\overline{AUC}_{\alpha}$} & \multicolumn{1}{c|}{\tiny $\overline{TPR}_{\alpha}$} & \multicolumn{1}{c|}{\tiny $\overline{FPR}_{\alpha}$} & \tiny{$\overline{AUC}_{\alpha}$} \\ \hline
		\multicolumn{1}{|c|}{0.069} & 3  & \multicolumn{1}{c|}{0.000} &  \multicolumn{1}{c|}{0.000} &0.533 &  \multicolumn{1}{c|}{0.667} & \multicolumn{1}{c|}{0.667}  &0.296 & \multicolumn{1}{c|}{0.758} & \multicolumn{1}{c|}{0.000} & 0.977 &  \multicolumn{1}{c|}{0.937} & \multicolumn{1}{c|}{0.032} & 0.987 &  \multicolumn{1}{c|}{0.952} & \multicolumn{1}{c|}{0.050} & 0.972 \\ \hline
		\multicolumn{1}{|c|}{0.096} & 3 & \multicolumn{1}{c|}{0.000} &  \multicolumn{1}{c|}{0.000} &0.528& \multicolumn{1}{c|}{0.333} & \multicolumn{1}{c|}{0.333} &0.595 &  \multicolumn{1}{c|}{0.959} & \multicolumn{1}{c|}{0.000} & 0.980 &  \multicolumn{1}{c|}{0.919} & \multicolumn{1}{c|}{0.067} & 0.973 &  \multicolumn{1}{c|}{0.951} & \multicolumn{1}{c|}{0.050} & 0.972 \\ \hline
		\multicolumn{1}{|c|}{0.134} & 3  & \multicolumn{1}{c|}{0.000} &  \multicolumn{1}{c|}{0.000} &0.610&  \multicolumn{1}{c|}{0.667} & \multicolumn{1}{c|}{0.667} & 0.539 & \multicolumn{1}{c|}{0.976} & \multicolumn{1}{c|}{0.000} & 0.988 &  \multicolumn{1}{c|}{0.961} & \multicolumn{1}{c|}{0.052} & 0.988 &  \multicolumn{1}{c|}{0.975} & \multicolumn{1}{c|}{0.050} & 0.987 \\ \hline
		\multicolumn{1}{|c|}{0.186} & 5 & \multicolumn{1}{c|}{0.384} & \multicolumn{1}{c|}{0.003} &0.746&  \multicolumn{1}{c|}{0.600} & \multicolumn{1}{c|}{0.600} &0.471 &   \multicolumn{1}{c|}{0.970} & \multicolumn{1}{c|}{0.000} & 0.985 &  \multicolumn{1}{c|}{0.973} & \multicolumn{1}{c|}{0.052} & 0.993 &  \multicolumn{1}{c|}{0.982} & \multicolumn{1}{c|}{0.050} &  0.991  \\ \hline
		\multicolumn{1}{|c|}{0.258} & 5 & \multicolumn{1}{c|}{0.929} & \multicolumn{1}{c|}{0.011} & 0.959 &  \multicolumn{1}{c|}{0.601} & \multicolumn{1}{c|}{0.644} &0.516 & \multicolumn{1}{c|}{0.988} & \multicolumn{1}{c|}{0.000} & 0.994  &  \multicolumn{1}{c|}{0.974} & \multicolumn{1}{c|}{0.057} & 0.996 &  \multicolumn{1}{c|}{0.994} & \multicolumn{1}{c|}{0.051} & 0.996  \\ \hline
		\multicolumn{1}{|c|}{0.359} & 5 & \multicolumn{1}{c|}{0.315} & \multicolumn{1}{c|}{0.000}  &0.975&  \multicolumn{1}{c|}{0.206} & \multicolumn{1}{c|}{0.213}  &0.437  & \multicolumn{1}{c|}{0.959} & \multicolumn{1}{c|}{0.000} & 0.979 &  \multicolumn{1}{c|}{0.991} & \multicolumn{1}{c|}{0.069} & 0.997 &  \multicolumn{1}{c|}{0.993} & \multicolumn{1}{c|}{0.050} &  0.996 \\ \hline
		\multicolumn{1}{|c|}{0.450} & 5 & \multicolumn{1}{c|}{0.000} &  \multicolumn{1}{c|}{0.000} &0.969&  \multicolumn{1}{c|}{0.729} & \multicolumn{1}{c|}{0.786}  &0.554 & \multicolumn{1}{c|}{0.978} & \multicolumn{1}{c|}{0.000} & 0.989  &  \multicolumn{1}{c|}{0.986} & \multicolumn{1}{c|}{0.061} & 0.996 & \multicolumn{1}{c|}{0.997} & \multicolumn{1}{c|}{0.050} & 0.998 \\ \hline
		\end{tabular}
		}
		\caption{3$\times$3 clean}
		 \label{tab:mnist_att3}
     \end{subtable}
     \label{tab:mnist}
      \vspace{-6mm}
\end{table*}

Table \ref{tab:mnist} shows the results obtained for different values of the poisoning ratio $\alpha$ for the three different attacks.
The values of $\overline{FPR}_{\alpha}(PC)$ and $\overline{TPR}_{\alpha}(PC)$ have been obtained by letting $\theta = \theta^*$.
%
For the clean-label case, due to the difficulty of developing a successful attack \cite{zhao2020clean, guo2023temporal, BarniKT19}, the backdoor can be successfully injected in the model only when $\alpha$ is large enough and, in any case, a successful attack could not always be obtained in the 10 repetitions.
The number of successfully attacked classes (cnt) with different poisoning ratios is reported in this case.
Upon inspection of Table \ref{tab:mnist}, we observe that:
\begin{itemize}
\item With regard to AC, the behaviour is similar under the three attack scenarios. Good results are achieved for intermediate values of $\alpha$, namely in the $[0.2, 0.3]$ range.  
However, AC can not defend against backdoor attacks adopting a poisoning ratio smaller than 0.1. Moreover, AC does not work when $\alpha> 0.5$, in which case $\overline{AUC}_{\alpha}$ goes to zero, as benign samples are judged as  poisoned and vice-versa (this is a consequence of the assumption made on the cluster size).
Finally, by comparing Table \ref{tab:mnist_att1} and  \ref{tab:mnist_att3}, we see that AC achieves better $\overline{AUC}_\alpha$ against the corrupted-label attack than in the clean-label case \BTcomm{This observation can be removed in case. Not so important.}.
\item With regard to CI, the detection performance achieved in the first attack scenario (3$\times$3 corrupted) is good for all the values of $\alpha$,
 (with the exception of the smallest $\alpha$, for which $\overline{AUC}_{\alpha} = 0.876$), showing  that  CI can effectively defend against the backdoor attack in this setting, for every attack poisoning ratio. However, as expected, CI fails in the other settings.
\item Regarding SCAn,
 when $\alpha<0.5$ results are very good in all the settings, with the only notable exception of small $\alpha$, in which case low  $\overline{AUC}_{\alpha}$ and, in particular, low $\overline{TPR}_{\alpha}(PC)$ values  are achieved.
 Similarly to AC, when $\alpha> 0.5$, $\overline{AUC}_{\alpha}$ goes to zero.
\item With regard to Beatrix,
 results are good especially in the first and third setting, a little bit worse in the second setting.
 Also in this case, the values of $\overline{TPR}_{\alpha}(PC)$ and $\overline{AUC}_{\alpha}$ are lower when $\alpha$ is small.

	\item CCA-UD provides good results in all settings and for every value of $\alpha$, with a perfect or nearly perfect $\overline{AUC}_{\alpha}$ in most of the cases. Moreover,
 a very good $\overline{TPR}_{\alpha}(PC)$ is obtained, \CH{always  larger (often much larger) than 0.95 with only very few exceptions in the corrupted attack settings when $\alpha$ is very low,  and a}  $\overline{FPR}_{\alpha}(PC)$ around 0.05.
 %
\end{itemize}

Overall, the tables confirm the universality of CCA-UD that works very well in all the setting, regardless of the specific attack  and regardless of the value of $\alpha$,
thus confirming that the  strategy used to detect poisoned clusters exploits a general misclassification behaviour.

\subsection{Results on traffic signs}
\label{sec:resultT}

\begin{figure*}
\centering
   \includegraphics[width=0.85\linewidth]{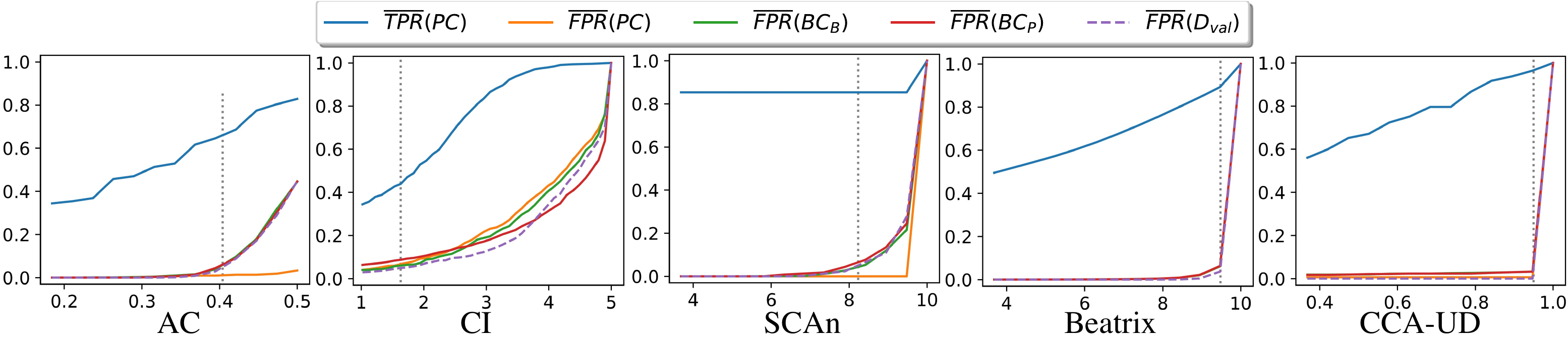}
\caption{Average  performance of AC, CI, SCAn, Beatrix, and CCA-UD for different values of $\theta$ for the traffic signs task. The vertical dotted line indicates the position of $\theta^*$ for the various methods. line. In each figure, the x-axis represents the threshold value, while the y-axis reports the values of $\overline{FPR}(BC_B)$, $\overline{FPR}(BC_P)$, $\overline{TPR}(PC)$ and $\overline{FPR}(PC)$ and $\overline{FPR}(D_{val})$. }
 \label{fig:exp4}
 \vspace{-3mm}
\end{figure*}

Fig.~\ref{fig:exp4} shows the average performance of AC, CI, SCAn, Beatrix, and CCA-UD on traffic signs.
Similar considerations to the MNIST case can be made.
CCA-UD achieves very good average performance at the operating point given by $\theta^*$, where $\overline{TPR}(PC)$ and $\overline{FPR}(PC)$ are (0.965, 0.058)
(with $\overline{FPR}(BC_B)=\overline{FPR}(BC_B)\approx 0.08$). As before,
for AC and CI a threshold that works well on the average can not be found.
In the case of a poisoned dataset, the average AUC of the detection $\overline{AUC}$ is equal to 0.897, 0.958, 0.924, 0.965, 0.993 for AC, CI, SCAn, Beatrix and CCA-UD, respectively.
We observe that also CI gets a good $\overline{AUC}_{\alpha}$. In fact,
given that the size of the input image is 28$\times$28, the trigger, namely the sinusoidal signal can be effectively removed by a 5$\times$5 average filter.

The results obtained for various $\alpha$ are reported in Table~\ref{tab:sm1_t}. As it can be seen, CCA-UD
achieves the best performance in terms of $\overline{AUC}_{\alpha}$,  and $\overline{TPR}_{\alpha}(PC)$ and $\overline{FPR}_{\alpha}(PC)$ measured at $\theta = \theta^*$, in all the cases.  
With regard to the other methods, as observed before, while CI and Beatrix are relatively insensitive to $\alpha$, the performance of AC and SCAn drop when $\alpha<0.1$ or $\alpha> 0.5$.
Also in this case, CCA-UD is the best-performing method outperforming  Beatrix (second-best), with a gain of about 0.03 in $\overline{AUC}_{\alpha}$ for every setting of $\alpha$, and  up to 0.12 in $\overline{TPR}_{\alpha}$.

\begin{figure*}
\centering
   \includegraphics[width=0.7\linewidth]{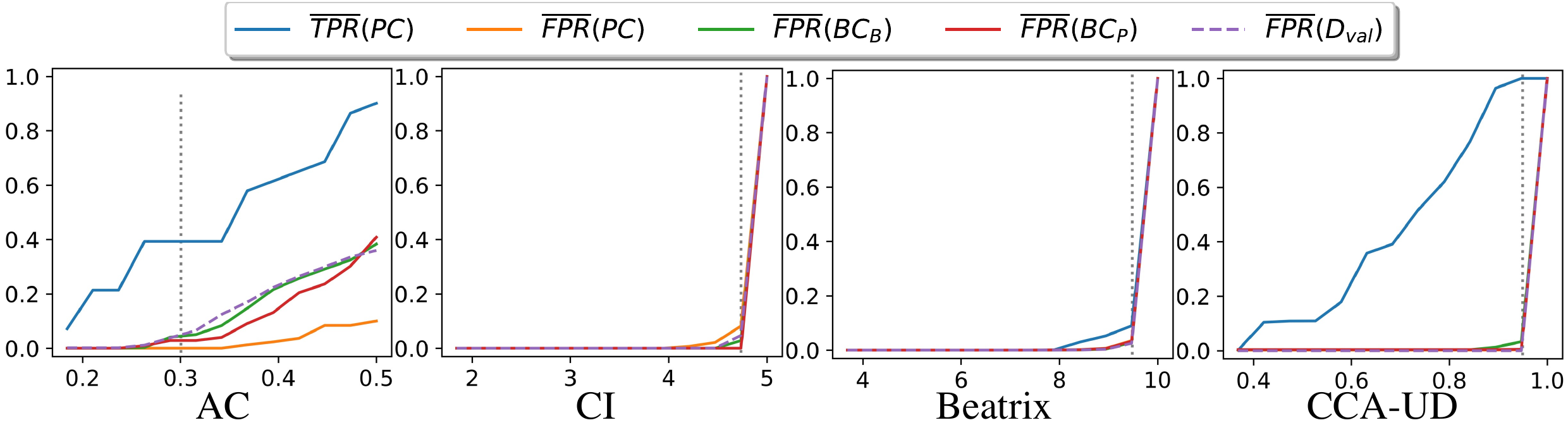}
\caption{Average  performance of AC, CI, Beatrix, and CCA-UD for different values of $\theta$ for the fashion clothes task. The vertical dotted line indicates the position of $\theta^*$ for the various methods. In each figure, the x-axis indicates threshold values, while the y-axis reports the values of $\overline{FPR}(BC_B)$, $\overline{FPR}(BC_P)$, $\overline{TPR}(PC)$ and $\overline{FPR}(PC)$ and $\overline{FPR}(D_{val})$}
   \label{fig:exp5}
 \vspace{-2mm}
\end{figure*}

\begin{table*}
\caption{Performance
for various poisoning ratios for the traffic signs task.  $\overline{FPR}_{\alpha}(PC)$ and $\overline{TPR}_{\alpha}(PC)$ (indicated as $\overline{FPR}_{\alpha}$ and $\overline{TPR}_{\alpha}$ in the table) values are computed with $\theta = \theta^*$.}
\centering
 \centering
\resizebox{1.7\columnwidth}{!}{
\begin{tabular}{cc|c|c|c|c|c|c|c|c|c|c|c|c|}
\cline{3-13}
&  & \multicolumn{1}{c|}{\textbf{{AC }}} & \multicolumn{1}{c|}{\textbf{{CI  }}} & \multicolumn{3}{c|}{\textbf{{SCAn}}} & \multicolumn{3}{c|}{\textbf{{Beatrix}}} & \multicolumn{3}{c|}{\textbf{{CCA-UD}}} \\ \hline
\multicolumn{1}{|c|}{$\alpha$} & cnt & \tiny{$\overline{AUC}_\alpha$} & \tiny{$\overline{AUC}_\alpha$}    & \tiny{$\overline{TPR}_{\alpha}$} & \tiny{$\overline{FPR}_{\alpha}$} & \tiny{$\overline{AUC}_\alpha$}  & \tiny{$\overline{TPR}_{\alpha}$} & \tiny{$\overline{FPR}_{\alpha}$}  & \tiny{$\overline{AUC}_\alpha$} &  \tiny{$\overline{TPR}_{\alpha}$} & \tiny{$\overline{FPR}_{\alpha}$} & \tiny{$\overline{AUC}_\alpha$} \\ \hline
\multicolumn{1}{|c|}{0.050} & 9 &  0.793 & 0.923 & \multicolumn{1}{c|}{0.728} & \multicolumn{1}{c|}{0.000} & 0.864  & \multicolumn{1}{c|}{0.929} & \multicolumn{1}{c|}{0.075}  & 0.973 & \multicolumn{1}{c|}{0.946} & \multicolumn{1}{c|}{ {0.061}} & {0.983}\\ \hline
\multicolumn{1}{|c|}{0.096} & 9 & 0.850& 0.928  & \multicolumn{1}{c|}{0.844} & \multicolumn{1}{c|}{0.000} & 0.922   & \multicolumn{1}{c|}{0.910} & \multicolumn{1}{c|}{0.079} & 0.973  & \multicolumn{1}{c|}{ {0.998}} & \multicolumn{1}{c|}{ {0.059}}  & {0.991}  \\ \hline
\multicolumn{1}{|c|}{0.134} & 9 & 0.949& 0.959  & \multicolumn{1}{c|}{0.884} & \multicolumn{1}{c|}{0.000} & 0.942   & \multicolumn{1}{c|}{0.878} & \multicolumn{1}{c|}{0.071} & 0.958  & \multicolumn{1}{c|}{ {0.998}} & \multicolumn{1}{c|}{ {0.057}} &  {0.992} \\ \hline
\multicolumn{1}{|c|}{0.186} & 10 & 0.958& 0.965 & \multicolumn{1}{c|}{0.867} & \multicolumn{1}{c|}{0.000} & 0.934  & \multicolumn{1}{c|}{0.873} & \multicolumn{1}{c|}{0.071} &  0.961 & \multicolumn{1}{c|}{0.999} & \multicolumn{1}{c|}{ {0.056}} &  {0.993}  \\ \hline
\multicolumn{1}{|c|}{0.359} & 13 & 0.946& 0.965  & \multicolumn{1}{c|}{0.925} & \multicolumn{1}{c|}{0.000} & 0.963  & \multicolumn{1}{c|}{0.897} & \multicolumn{1}{c|}{0.079}  & 0.960  & \multicolumn{1}{c|}{ {0.985}} & \multicolumn{1}{c|}{ {0.054}} & {0.996} \\ \hline
\multicolumn{1}{|c|}{0.450} & 14 & 0.917& 0.965  & \multicolumn{1}{c|}{0.908} & \multicolumn{1}{c|}{0.000} & 0.954   & \multicolumn{1}{c|}{0.915} & \multicolumn{1}{c|}{0.070}& 0.963& \multicolumn{1}{c|}{ {0.980}} & \multicolumn{1}{c|}{ {0.055}}  & {0.994} \\ \hline
\multicolumn{1}{|c|}{0.550} & 15 & 0.869 & 0.996 & \multicolumn{1}{c|}{0.782} & \multicolumn{1}{c|}{0.000} & 0.891 & \multicolumn{1}{c|}{0.934} & \multicolumn{1}{c|}{0.085} & 0.971  & \multicolumn{1}{c|}{0.999}  & \multicolumn{1}{c|}{ {0.051}}  & {0.999} \\ \hline
\end{tabular}
}
\label{tab:sm1_t}
\vspace{-3mm}
\end{table*}

\begin{table}
\caption{Performance
for various poisoning ratios for the fashion clothes task.  $\overline{FPR}_{\alpha}(PC)$ and $\overline{TPR}_{\alpha}(PC)$ (indicated as $\overline{FPR}_{\alpha}$ and $\overline{TPR}_{\alpha}$) values are computed with $\theta = \theta^*$.}
 \centering
\resizebox{1\columnwidth}{!}{
\begin{tabular}{cc|c|c|c|c|c|c|c|c|c|c|}
\cline{3-10}
&  & \multicolumn{1}{c|}{\textbf{{AC  }}} & \multicolumn{1}{c|}{\textbf{{CI }}} & \multicolumn{3}{c|}{\textbf{{Beatrix }}} & \multicolumn{3}{c|}{\textbf{{CCA-UD}}} \\ \hline
\multicolumn{1}{|c|}{$\alpha$} & cnt & \tiny{$\overline{AUC}_\alpha$} & \tiny{$\overline{AUC}_\alpha$} & \tiny{$\overline{TPR}_{\alpha}$} & \tiny{$\overline{FPR}_{\alpha}$}  &  \tiny{$\overline{AUC}_\alpha$}  & \tiny{$\overline{TPR}_{\alpha}$} & \tiny{$\overline{FPR}_{\alpha}$}  & \tiny{$\overline{AUC}_\alpha$} \\ \hline
\multicolumn{1}{|c|}{0.069} & 3 & 0.618 & 0.056  & \multicolumn{1}{c|}{0.113} & \multicolumn{1}{c|}{0.059} & 0.546  & 1.000 & 0.052 & 0.998 \\ \hline
\multicolumn{1}{|c|}{0.096} & 3 & 0.513 & 0.341  & \multicolumn{1}{c|}{0.025} & \multicolumn{1}{c|}{0.047} & 0.456   & 1.000 & 0.056 & {0.995} \\ \hline
\multicolumn{1}{|c|}{0.134} & 3 & 0.940 & 0.087  & \multicolumn{1}{c|}{0.025} & \multicolumn{1}{c|}{0.074} & 0.508  & 1.000 & 0.053 & 0.998  \\ \hline
\multicolumn{1}{|c|}{0.186} & 4 & 1.000 & 0.037  & \multicolumn{1}{c|}{0.032} & \multicolumn{1}{c|}{0.059} & 0.425    & 1.000 & 0.055 & {0.998} \\ \hline
\multicolumn{1}{|c|}{0.258} & 5 & 1.000 & 0.083 & \multicolumn{1}{c|}{0.026} & \multicolumn{1}{c|}{0.061} & 0.413 & 1.000 &  {0.057}  & {0.996}  \\ \hline
\multicolumn{1}{|c|}{0.359} & 5 & 1.000 & 0.015 & \multicolumn{1}{c|}{0.300} & \multicolumn{1}{c|}{0.060} & 0.697   & 1.000 &  {0.052}& {0.998} \\ \hline
\multicolumn{1}{|c|}{0.450} & 5 & 1.000 & 0.174 & \multicolumn{1}{c|}{0.207} & \multicolumn{1}{c|}{0.059} & 0.591  & 1.000 &  {0.050} & {1.000}  \\ \hline
\end{tabular}
}
\vspace{-6mm}
\label{tab:sm2_t}
\end{table}

\subsection{Results on fashion clothes}
\label{sec:resultF}

Fig.~\ref{fig:exp5} reports the results obtained on the fashion clothes task. We did not run the SCAn method in this case. In fact, as already pointed in \cite{MaWSXWX23}, SCAn is highly time-consuming \BTcomm{is the issue the time or the resources?} 
when the feature dimension is large. With this network $d = 9216$ (with MNIST and traffic sign $d$ is lower than 512), SCAn took more than 7 days to find out the poisoned samples from the training dataset, running on a computer with Intel(R) Core(TM) i7-8700 CPU@3.20GHz.

Once again, the performance of CCA-UD is largely superior to those achieved by other works. In particular, by looking at Fig.~\ref{fig:exp5}, CCA-UD achieves $\overline{TPR}(PC)$ and $\overline{FPR}(PC)$ equal to (1.000,  {0.053}), with $\overline{FPR}(BC_B)=\overline{FPR}(BC_P)\approx 0.05$.
Regarding the AUC scores, $\overline{AUC}$ of AC, CI, Beatrix, and CCA-UD are 0.900, 0.106, 0.519 and 0.997 respectively.
%
Since the attack is carried out in a clean-label modality,
the poor performance of CI is expected.
%
The bad performance of Beatrix in this case, instead, is likely due to the curse of dimensionality,
\CH{since distinguishing between poisoned samples and benign ones by relying on  inner  products of feature vectors
is not possible when the feature dimension is very large (as it is the case here, with a dimension larger than 9000)}. 
%
The results for various  $\alpha$ are reported in Table \ref{tab:sm2_t} and confirm the good behaviour of CCA-UD, which provides very good performance in all the cases,
always outperforming the other  methods.



\subsection{Other datasets, architectures and attacks}
\label{sec.others}


Below we report the additional experiments we carried out with different and more complex tasks and attacks, to prove the generality/universality of the proposed method.

\subsubsection{CIFAR10 classification}
CIFAR10 dataset 
consists of a total of 60000 images of size  32$\times$32 belonging to 10 classes,
split into two parts (50000 for training and 10000 for testing).
The model architecture is based on VGG19 \BTcomm{AS I said, I have removed the references to known DL architectures and mathematical  methods used in other papers. WE can reinsert them if in some cases we believe they are necessary.}, 
 and the feature representation is extracted from the final convolutional (16th) layer after the pooling layer and flatten operation. The following types of corrupted-label backdoor attacks are considered:
\begin{itemize}
    \item Corrupted-label attack, with 3$\times$3 pixel trigger, see Fig.~\ref{fig:trigger}. Specifically, the attacker chooses samples from non-target classes, adds the trigger over the samples, and finally modifies the labels to the target class.
    \item Source-specific attack, with 3$\times$3 pixel trigger, see Fig.~\ref{fig:trigger}. The attacker poisons  samples from a specific source class and modifies their labels into that of the target class. At test time, only poisoned samples from this class can lead to misclassification.
    \item Sample-specific attack. The attack is carried out considering the warping-based trigger described in~\cite{NguyenT21}, see Fig.~\ref{fig:trigger}.
 The poisoned images are warped with fixed parameters and the labels modified into that of the target class.
 To facilitate the network to learn the specific backdoor, the attacker also injects into the training dataset noise samples, namely images warped randomly,
 for which the labels are not corrupted.
\end{itemize}


In this case, to speed up the experiments, we run our tests by randomly choosing the target class, instead of repeating the experiments for every possible choice of the target. Experiments are carried out considering 5 different
poisoning ratios, ranging from 0.096 to 0.45. \BTcomm{@David: Now I do not understand / remember  why for some tasks we stopped at 0.45 while for others we go above 0.5 and reach 0.55, where some methods from sota fail while our method works well.} 

\begin{figure}
\begin{subfigure}[h]{0.5\textwidth}
\centering
   \includegraphics[width=0.9\linewidth]{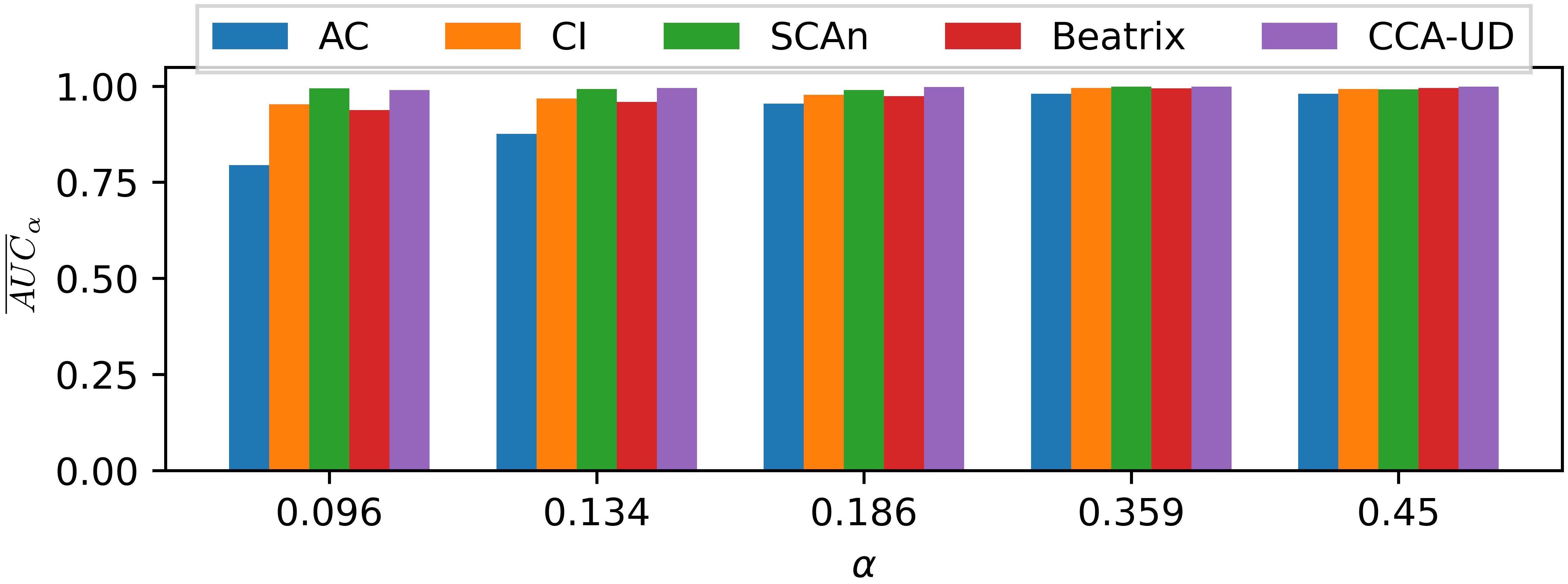}
   \vspace{-2mm}
   \caption{Corrupted-label}
   \label{fig:corrupted}
\end{subfigure}\\
\begin{subfigure}[h]{0.5\textwidth}
\centering
   \includegraphics[width=0.9\linewidth]{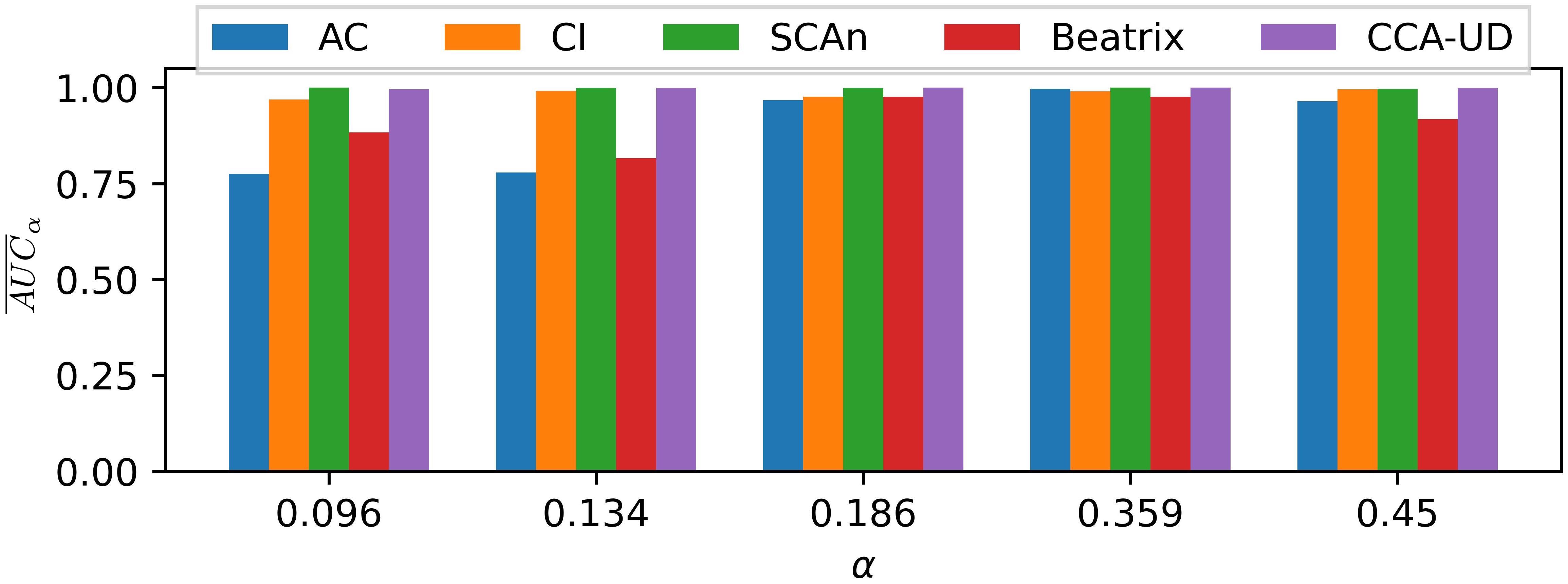}
   \vspace{-2mm}
   \caption{Source-specific}
   \label{fig:source_specific}
\end{subfigure}\\
\begin{subfigure}[h]{0.5\textwidth}
\centering
   \includegraphics[width=0.9\linewidth]{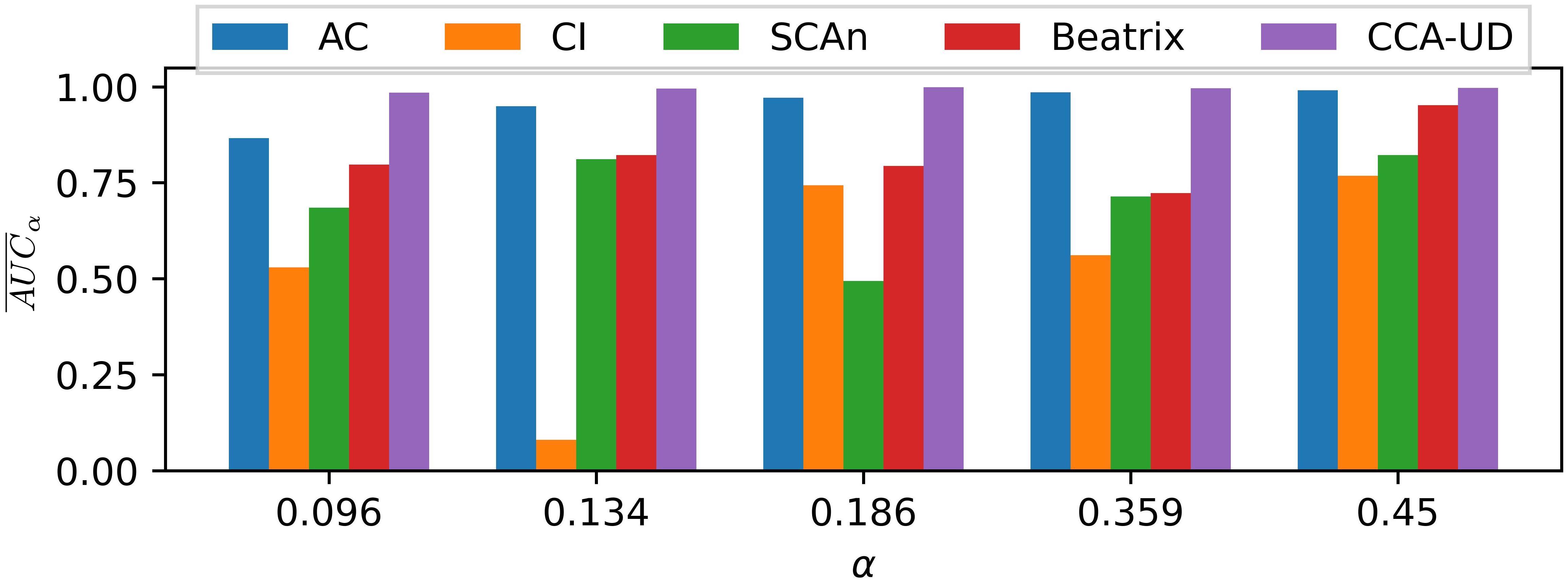}
    \vspace{-2mm}
   \caption{Sample-specific}
   \label{fig:sample_specifc}
\end{subfigure}
\caption{Performance ($\overline{AUC}_\alpha$) of AC, CI, SCAn, Beatrix and CCA-UD for various poisoning ratio $\alpha$ for CIFAR10 classification, against the three types of backdoor attacks. \BTcomm{Strictly speaking, the y-axis should report 'AUC' with the correct formalism used in the paper. Same after}
}
 \vspace{-4mm}
\end{figure}

The $\overline{AUC}_\alpha$ obtained for the three types of attack are shown in Fig. \ref{fig:corrupted}, Fig. \ref{fig:source_specific}, and Fig. \ref{fig:sample_specifc}, respectively.
We verified that SCAn, Beatrix, and CCA-UD\footnote{For AC and CI, it is hard to find a fixed threshold, as discussed in Section \ref{sec.thresholdresults}, so we only compare CCA-UD with SCAn and Beatrix.} can achieve average $FPR$, evaluated on the benign class, as $0.066, 0.052$, and $0.003$, respectively. \BTcomm{@David: we should say something at least in the text on the false positive rate achieved by the various method, e.g. that we checked that it is around 0.05 in all the cases.}

For the corrupted-label attack in Fig. \ref{fig:corrupted}, we can observe that CI, SCAn, Beatrix and CCA-UD can achieve good performance for different poisoning ratios with $\overline{AUC}_\alpha\approx 1$. However, AC performance degrades to 0.77 when the poisoning ratio decreases.

In the case of the source-specific attack shown in Fig. \ref{fig:source_specific}, all methods work very well when $\alpha\ge 0.186$. AC and Beatrix are the methods showing the worse performance (with an $\overline{AUC}_\alpha\approx 0.75$) when $\alpha$ is small ($\alpha\leq 0.134$ in the plots).

By choosing the threshold as explained in Section \ref{sec:threshold}, we were able to find a  threshold $\theta^*$
that works in all cases, also
for this task.
Specifically, by choosing the threshold in this way (set to {9.74, 9.38, 0.98}) we get
$\overline{TPR}(PC)$ and $\overline{FPR}(PC)$ equal to (0.775, 0.001), (0.671, 0.078), (0.996, 0.002) for different $\alpha$'s and three different tasks. \BTcomm{@David: I guess that you are referering to the average values right (that is, $\overline{TPR}(PC)$ and not $\overline{TPR}_{\alpha}(PC)$ - same for the others) ?  can you be more precise for FPR ? What does $\approx 0$ measn? isn't it approx 0.05 as before?  }.  

From the results obtained against the sample-specific attack shown in Fig. \ref{fig:sample_specifc}, we can observe that  CCA-UD  achieves the best performance with $\overline{AUC}_\alpha\approx 1$ in all the cases. With regard to the other methods, we can observe
the following:
1) AC achieves good results when $\alpha$ is large (being always smaller than 0.5), while - as before - performance drops when $\alpha$ becomes very small;
2) CI is not very effective, since the average filter cannot remove the warping-based trigger; 3) SCAn's bad performance was expected since, as observed in \cite{MaWSXWX23}, this method can not work on sample-specific attacks 
Beatrix can indeed improve the performance of SCAn in this case, however, performance is good only when $\alpha$ is large.



\subsubsection{Face recognition task}

We also run some experiments considering a very different task, namely, face recognition,  with larger-size images. For this task, features tend to have a larger intra-class variability compared to the other tasks. To prove this, we visualise the feature distribution for this task in Section III of the supplementary material and compare it with CIFAR10. \BTcomm{Is this true? CAn we say this? having a picture in Fig \ref{fig:intra_var} that shows the lower variability for the CIFAR10 case would be nice.} 

For these experiments, the 12 most populated classes in YoutubeFace dataset \cite{WolfHM11} were selected, with more than 2600 images per class. The  dataset is split in  training and testing in proportions 9:1. The image size is 315$\times$315. Classification is performed considering an Inception-Resnet-v1 architecture \cite{SzegedyIVA17}. The feature representation for the clustering analysis  are extracted before the first FC layer (2nd last layer).




With regard to the attack, we considered the corrupted label case, with a 30$\times$30 pixel trigger, see Type1 trigger in Fig. \ref{fig:trigger}. The same triggering signal used before was considered,  enlarged by a factor of 10 (being the size of the images approximately 10 times larger than in the previous cases, this triggering signal has approximately the same relative size as before).

As for the case of CIFAR10,  we run tests for a randomly chosen target class.
Experiments were carried out by considering 5 different poisoning
ratios, i.e., 0.05, 0.134, 0.359, 0.45, 0.55.

\begin{figure}
\centering
   \includegraphics[width=0.9\linewidth]{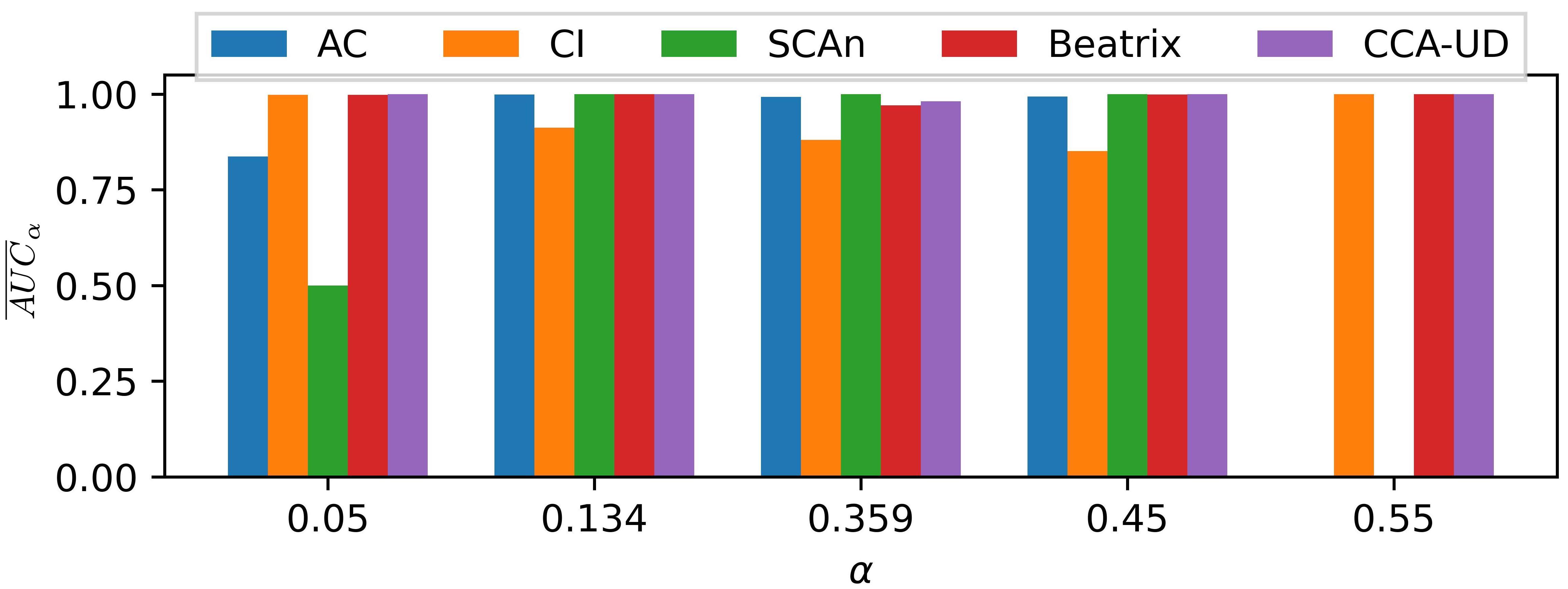}
    \vspace{-2mm}
   \caption{Performance ($\overline{AUC}_\alpha$) of AC, CI, SCAn, Beatrix and CCA-UD for various poisoning ratio $\alpha$ for face recognition (YouTubeFace), against the 30$\times$30 pixel trigger attacks.
   }
   \label{fig:face_rec}
 \vspace{-4mm}
\end{figure}

Fig. \ref{fig:face_rec} shows the $\overline{AUC}_\alpha$ values. Performance is good for most of the methods. CCA-UD and Beatrix are the best-performing methods, with an almost perfect $\overline{AUC}_\alpha$ in all the settings.
With regard to the other methods, a behaviour similar to the previous cases can be observed. In particular,  AC and SCAn do not work when $\alpha=0.55$, and their performance drops (especially for SCAn) when $\alpha$ is very small (cluster imbalance issue). CI is the method that has worse performance on average, however, some discrimination capability can be observed also for this method. In fact, even if the average filter kernel is 5$\times$5 (and the filter can not completely remove the trigger), the triggering signal is impaired by the filter and the activation of the backdoor inhibited.

For $FPR$, we evaluated the performance of SCAn, Beatrix and CCA-UD on the benign class and got average values equal to 0.086, 0.053, and 0.002. Given the thresholds (8.46, 9.65, and 0.95 for SCAn, Beatrix and CCA-UD) determined as in Section \ref{sec.thresholdresults}, the average $\overline{TPR}(PC)$ and $\overline{FPR}(PC)$ are equal to (0.800, 0.100), (0.961, 0.053), (0.988, 0.000), respectively.

In summary, these results confirm that CCA-UD is effective also when there is large intra-variability in the feature representation of the various classes and then the samples in benign classes are split  into distinct clusters (as shown in Section III of supplementary material). Being these clusters benign, as expected, the clusters' centroids do not activate the misclassification behaviour CCA-UD looks for.


For the face recognition task, we also run an experiment considering a multiple triggers attack,  to evaluate the effectiveness of CCA-UD also in this scenario. In a multiple triggers attack, several triggers are used to poison the samples,
to induce more than one malicious behaviour. Specifically, in our experiments, the attacker chooses three types of 30$\times$30 triggers to poison three different classes, as shown in Fig. \ref{fig:trigger}, where the three triggers use different patterns and are placed in different locations. At test time, the presence of the triggering signal inside the sample will lead to a misclassification in favour of the corresponding target class.

Our experiments confirm that CCA-UD can achieve very good performance, with an average $\overline{AUC}\approx 0.99$,
for all the target classes. 
At the optimum  threshold $\theta^*=0.95$ (this threshold is the same as for the previous experiment as it was set on benign data), we get an error probability averaged over the three classes equal to $\overline{TPR}(PC) = 0.997$ and $\overline{FPR}(PC) = 0.006$.

On the same experiments SCAn and Beatrix achieved $\overline{AUC} = 0.833$ and  $\overline{AUC} = 0.999$ respectively, and identified the poisoned samples  with  $\overline{TPR}(PC)$ and $\overline{FPR}(PC)$ equal to (0.667, 0.001) and (0.998, 0.064), respectively.
\BTcomm{The description and the numbers can not be exactly the same for this attack and the previous attack (see also my previous comment). CAn you be more precise and give precise numbers? However, it makes sense that the threshold is the same (I stressed this). Also, since you are speaking about average I guess that here you are reporting the metrics without the pedex.} 

\section{Concluding remarks}
\label{sec:conclude}

We have proposed a universal backdoor detection method, called CCA-UD,
to reveal the possible presence of a backdoor inside a model and identify the poisoned samples by analysing the training dataset. CCA-UD relies on DBSCAN clustering and on a new strategy for the detection of poisoned clusters based on the analysis of clusters' centroids, that exploits a general behaviour of backdoored models. The capability of the centroids' features to cause misclassification of benign samples is exploited to decide whether a cluster is poisoned or not.
We evaluated the effectiveness of CCA-UD on a great variety of classification tasks and architectures, and attack scenarios. The results confirm that CCA-UD can work well regardless of the corruption strategy (corrupted- or clean-label), the poisoning ratio (that can either be very small or very large), and the type of trigger used by the attacker.
In particular, in our experiments, we considered a wide variety of triggers, from fixed triggering signals (local and global pattern) to source-specific and sample-specific triggers.
Furthermore, we proved that the performance achieved by CCA-UD is  always superior or comparable to those achieved by the existing methods, when these methods are applied in a setting that meets their operating requirements.

Future work will be devoted
to the investigation of the capability of CCA-UD to defend against backdoor attacks in application scenarios beyond image classification,
The effectiveness of the method against backdoor attacks carried out against  modern architectures (e.g. vision transformers) \CH{or different deep neural network models, e.g. generative models \cite{ChouCH23}, sequential models \cite{nallapati2017summarunner}, and recurrent neural networks in particular,}  is also worth investigation.


\bibliographystyle{IEEEtran}
\bibliography{ref.bib}

\end{document}